\documentclass[journal,twoside]{IEEEtran}
\usepackage[pdftex]{graphicx}
\DeclareGraphicsExtensions{.pdf,.jpg,.png}
\usepackage{amsmath,amssymb}
\usepackage{algorithm}
\usepackage{algorithmic}
\usepackage{array}
\usepackage{color}
\usepackage[pagebackref=true,breaklinks=true,colorlinks,bookmarks=false,linkcolor=black,anchorcolor=black,citecolor=black, urlcolor=black]{hyperref}
\usepackage{bm}
\usepackage{adjustbox}
\usepackage{multirow}
\usepackage{subfigure}
\usepackage{threeparttable}
\usepackage{enumitem}
\usepackage{makecell}
\usepackage{booktabs}
\usepackage{cite}
\usepackage{tikz}
\usepackage{float}
\usepackage{soul}
\soulregister\cite7 
\soulregister\ref7 
\soulregister\pageref7

\newcommand{\orcidicon}{\includegraphics[width=0.32cm]{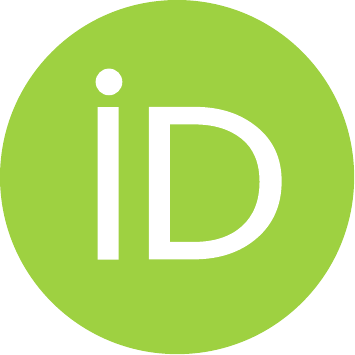}}
\foreach \x in {A, ..., Z}{
	\expandafter\xdef\csname orcid\x\endcsname{\noexpand\href{https://orcid.org/\csname orcidauthor\x\endcsname}{\noexpand\orcidicon}}
}

\begin{document}

%
\title{Double Similarity Distillation for Semantic Image Segmentation}
%
%
%

\author{Yingchao~Feng\orcidA{},~\IEEEmembership{Graduate Student Member,~IEEE}, Xian~Sun\orcidC{},~\IEEEmembership{Senior Member,~IEEE}, Wenhui~Diao\orcidB{},~\IEEEmembership{Member,~IEEE,} Jihao~Li,~\IEEEmembership{Graduate Student Member,~IEEE}, and~Xin~Gao%
\thanks{
	This work was supported by the National Natural Science Foundation of China under Grant 41701508. \emph{(Corresponding author: Xian Sun.)}}
\thanks{
	Y. Feng, X. Sun and J. Li are with the Aerospace Information Research Institute, Chinese Academy of Sciences, Beijing 100190, China, also with the School of Electronic, Electrical and Communication Engineering, University of Chinese Academy of Sciences, Beijing 100190, China, also with the University of Chinese Academy of Science, Beijing 100190, and also with the Key Laboratory of Network Information System Technology (NIST), Aerospace Information Research Institute, Chinese Academy of Sciences, Beijing 100190, China (e-mail: fengyingchao17@mails.ucas.edu.cn; sunxian@aircas.ac.cn; lijihao17@mails.ucas.edu.cn).}
\thanks{
	W. Diao and X. Gao are with Aerospace Information Research Institute, Chinese Academy of Sciences, Beijing 100190, China, and also with the Key Laboratory of Network Information System Technology (NIST), Aerospace Information Research Institute, Chinese Academy of Sciences, Beijing 100094, China (e-mail: diaowh@aircas.ac.cn; gaoxin@aircas.ac.cn).}}



\markboth{IEEE TRANSACTIONS ON IMAGE PROCESSING}%
{Feng \MakeLowercase{\textit{et al.}}: Double Similarity Distillation for Semantic Image Segmentation}

%



\maketitle



\begin{abstract}
The balance between high accuracy and high speed has always been a challenging task in semantic image segmentation. Compact segmentation networks are more widely used in the case of limited resources, while their performances are constrained. In this paper, motivated by the residual learning and global aggregation, we propose a simple yet general and effective knowledge distillation framework called double similarity distillation (DSD) to improve the classification accuracy of all existing compact networks by capturing the similarity knowledge in pixel and category dimensions, respectively. Specifically, we propose a pixel-wise similarity distillation (PSD) module that utilizes residual attention maps to capture more detailed spatial dependencies across multiple layers. Compared with exiting methods, the PSD module greatly reduces the amount of calculation and is easy to expand. Furthermore, considering the differences in characteristics between semantic segmentation task and other computer vision tasks, we propose a category-wise similarity distillation (CSD) module, which can help the compact segmentation network strengthen the global category correlation by constructing the correlation matrix. Combining these two modules, DSD framework has no extra parameters and only a minimal increase in FLOPs. Extensive experiments on four challenging datasets, including Cityscapes, CamVid, ADE20K, and Pascal VOC 2012, show that DSD outperforms current state-of-the-art methods, proving its effectiveness and generality. The code and models will be publicly available.
\end{abstract}


%
\IEEEpeerreviewmaketitle

\section{Introduction}
\label{sec:intro}

\begin{figure}[t]
	\centering
	\includegraphics[width=\linewidth]{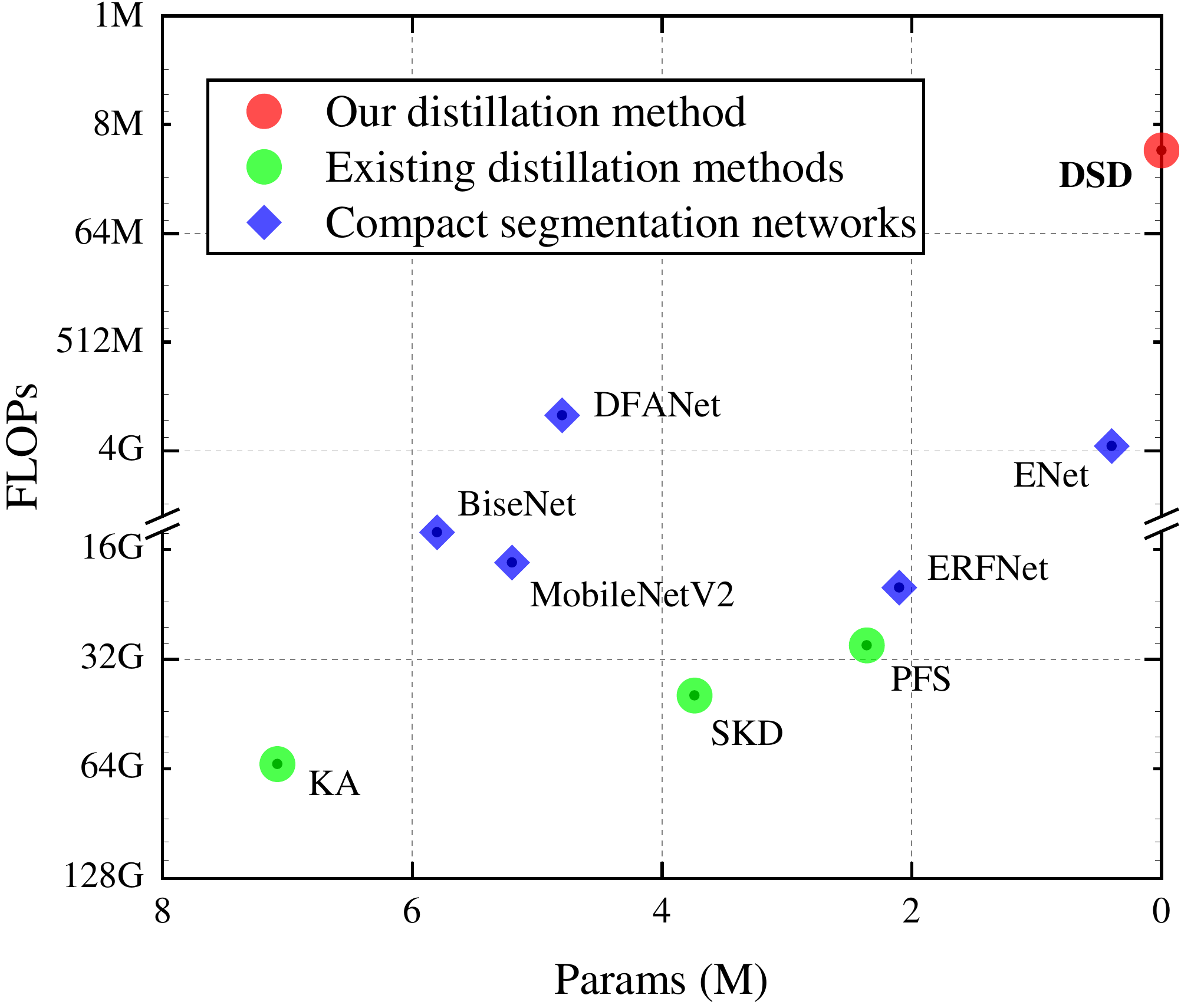}
	\caption{FLOPs and parameters of different distillation methods for semantic image segmentation. The FLOPs is calculated with the resolution of $3 \times 640 \times 360$. The red circle is our proposed distillation method, while green circles are the existing distillation methods KA~\cite{He_2019_CVPR}, SKD~\cite{Liu_2019_CVPR,liu2020structured}, and PFS~\cite{shan2019distilling}. The blue diamonds are the recently proposed compact segmentation networks, including MobileNetV2~\cite{Sandler_2018_CVPR}, ERFNet~\cite{romera2017erfnet}, DFANet~\cite{Li_2019_CVPR}, ENet~\cite{paszke2016enet}, BiseNet~\cite{Yu_2018_ECCV}. The existing distillation methods need high FLOPs and introduce a lot of parameters, which even exceed those of the compact networks. On the contrary, our proposed distillation method has no extra parameters and only a minimal increase in FLOPs.}
	\label{fig:com_param_flops}
\end{figure}

\IEEEPARstart{C}{urrent} state-of-the-art segmentation networks~\cite{zhao2017pyramid,Zhao_2018_ECCV,Yang_2018_CVPR,ding2020semantic,ding2018context} are not suitable for real-time applications with low power consumption and low storage, such as autonomous driving, augmented/virtual reality, and video surveillance, especially when such applications work on smartphones, AR/VR devices, and other edge devices with strict constraints on energy, memory, and computation. These methods are hard to deploy due to their high computational complexity and large model size. This has led to a new research field, which focuses on the design of compact segmentation networks.

Approaches in this field are broadly divided into three categories. The first category is the simplest way to achieve fast segmentation by directly compressing the cumbersome models, such as weight pruning~\cite{Lebedev_2016_CVPR}, weight decomposition~\cite{zhang2015accelerating,NIPS2015_5787}, and weight quantization~\cite{Rastegari_xnor7,Jacob_2018_CVPR}. The second category replaces the large and heavy backbone networks with compact models or thinner and shallower models. For example, using MobileNetV2~\cite{Sandler_2018_CVPR} instead of ResNet-152~\cite{he2016deep} or reducing the depth and scaling the width (e.g., ResNet-18). Most works belong to the third category, which is to design specific compact segmentation networks, such as ICNet~\cite{Zhao_2018_ECCV_icnet}, ERFNet~\cite{romera2017erfnet}, and DFANet~\cite{Li_2019_CVPR}. However, there still exists a certain margin between the performance of these compact networks and the cumbersome networks. The compact networks always produce poor prediction accuracy on boundaries, small targets, and weak categories.

How to improve the performance of compact semantic image segmentation networks has received increased attention. The knowledge distillation strategy~\cite{hinton2015distilling} has been investigated as a good alternative, which is a popular solution to improve the performance of compact models (also called student networks) by transferring knowledge from the cumbersome models (also called teacher networks). The effectiveness of this solution has been verified in several fields, such as image classification~\cite{adriana2015fitnets,Zagoruyko2017AT}, face recognition~\cite{luo2016face}, object detection~\cite{Wang_2019_CVPR}, lane detection~\cite{hou2019learning}, degraded image segmentation~\cite{guo2019degraded} and continual learning~\cite{8107520}. Recently, some of the work including MDE~\cite{xie2018improving}, KA~\cite{He_2019_CVPR}, SKD~\cite{Liu_2019_CVPR,liu2020structured}, and PFS~\cite{shan2019distilling}, introduce the knowledge distillation strategy into the segmentation field, and indicate that spatial dependencies play a powerful role in improving the performance of the compact network. However, there are still serious obstacles to the applications in the realistic environment.

\begin{figure}[t]
	\centering
	\subfigcapskip=5pt
	\subfigure{\begin{minipage}{.98\linewidth}
			\label{fig:ps}
			\centering
			\includegraphics[width=\linewidth]{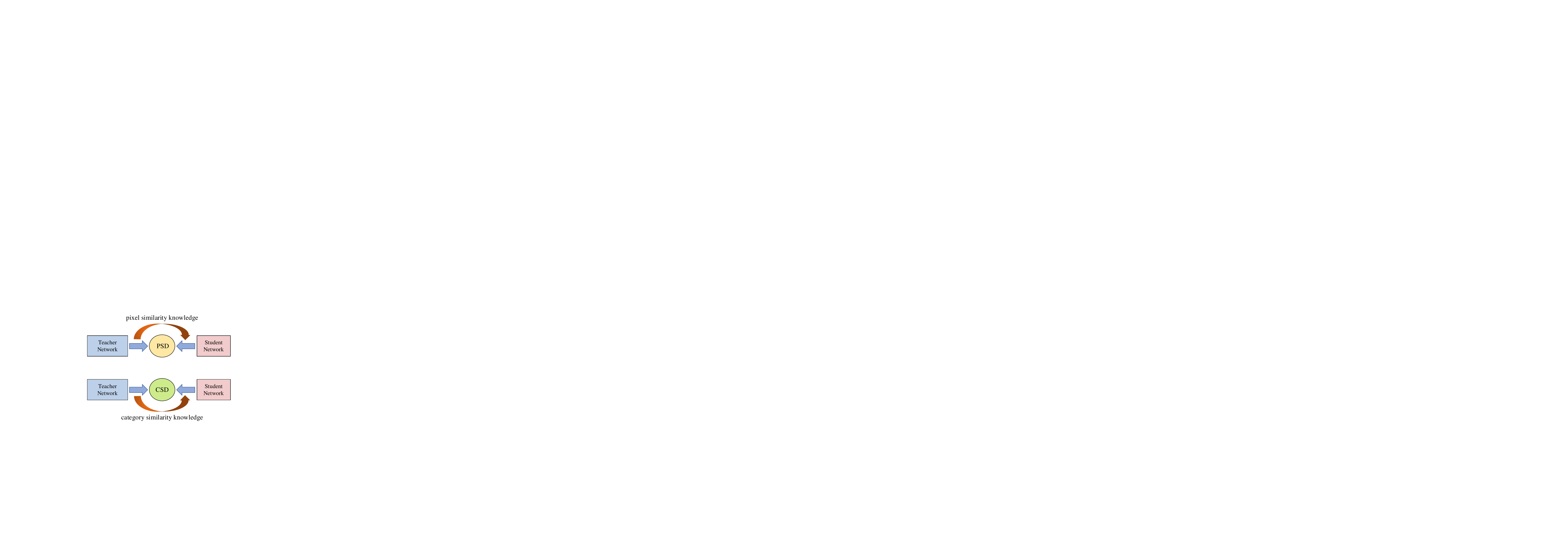}
	\end{minipage}}
	\vfill
	\subfigure{\begin{minipage}{.98\linewidth}
			\label{fig:cs}
			\centering
			\includegraphics[width=\linewidth]{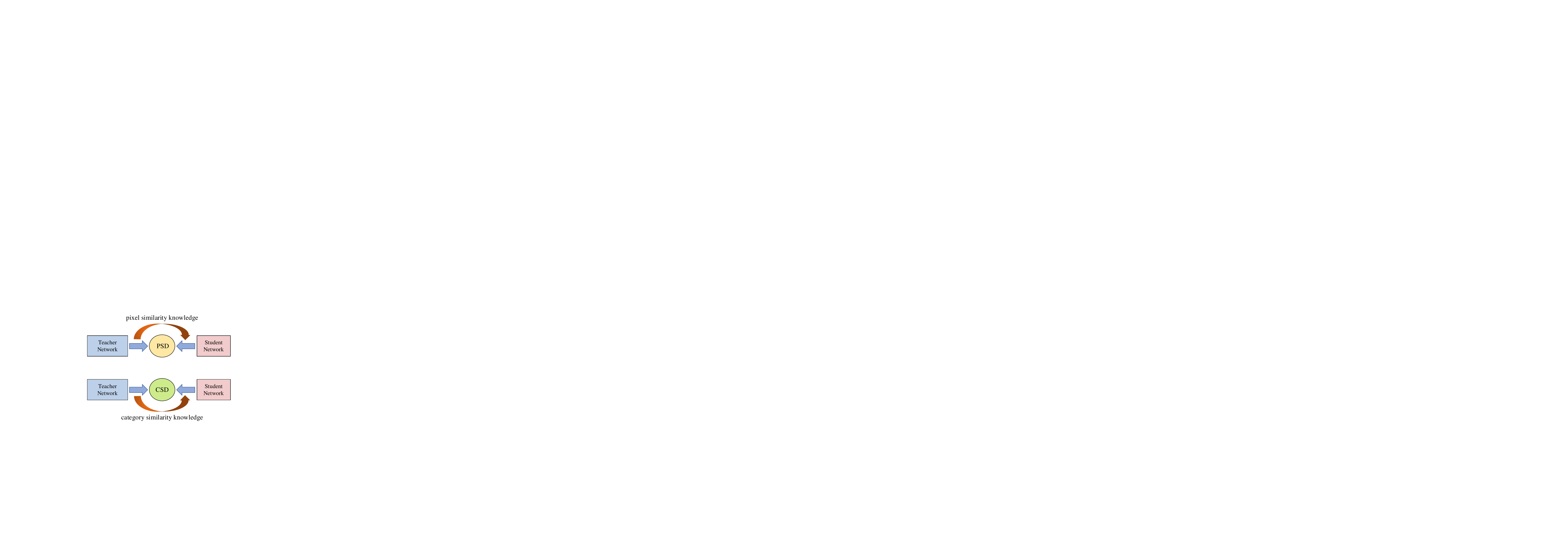}
	\end{minipage}}
	\caption{Knowledge transferred from our proposed double similarity distillation framework. The PSD module is applied between the teacher network and the student network to capture detailed spatial dependencies. The CSD module helps the student network to extract the global category correlation.}
	\label{fig:dual sim}
\end{figure}

As noted by the Occam's razor, ``Entities should not be multiplied unnecessarily.'' For the existing methods, the major problem is the high complexity and extra parameters caused by a large number of matrix multiplication and network structures that need to be optimized. Specifically, it can be summarized in two parts. Firstly, the existing methods utilize the matrix multiplication on a three-dimensional feature map to capture the spatial dependencies, the high complexity ($\mathcal{O}(n^2)$, $n$ means product of the spatial dimensions of the corresponding feature map) makes it applicable only at the output layer of the network, which limits the performance gains. Secondly, the extra network structures, such as auto-encoder network and generative adversarial network, are introduced to make up for the deficiency. Although the complexity is reduced to some extent, these network structures lead to increase the number of parameters and training steps. As shown in Fig.~\ref{fig:com_param_flops}, the existing distillation methods need high FLOPs and introduce a great number of parameters, which even exceed those of the common compact segmentation networks. The high complexity and extra parameters greatly limit their scope of application and practicality. Besides, some of them simply take the segmentation task as many separate pixel classification tasks and directly adopt pixel-based category correlation, which only achieves sub-optimal improvements.

In this paper, we propose a novel general method, called double similarity distillation (DSD). We aim to build a simple and effective knowledge distillation framework to transfer comprehensive and powerful similarity knowledge from the cumbersome teacher network to the compact student network. As shown in Fig.~\ref{fig:dual sim}, the similarity knowledge in pixel and category dimensions are introduced in DSD. The pixel-wise similarity distillation (PSD) module captures the detailed spatial dependencies and the category-wise similarity distillation (CSD) module extracts the global category correlation. 

In contrast with other methods that capture the spatial dependencies with high complexity, the PSD module utilizes residual attention maps through subtraction between any two attention maps. Compared with matrix multiplication on feature maps, the lower complexity ($\mathcal{O}(n)$ vs $\mathcal{O}(n^2)$) allows the PSD module to be applied to multiple layers of the network simultaneously to capture more detailed spatial dependencies. The behind intuitions come from the residual learning~\cite{he2016deep} and attention transfer~\cite{Zagoruyko2017AT}: The responses of pixels belonging to the same category on the residual attention map should be similar. 
Furthermore, we also propose the CSD module to strengthen the global category correlation through the category correlation matrix, which can be complementary to the PSD module. For one image, due to complex neighborhood information, there are huge differences among the prediction distributions of pixels belonging to the same category. Directly applying the soft target distillation for each pixel individually to learn the pixel-based category correlation may contain a lot of noise information. This increases the learning difficulty of the student network. Based on the perspective of global aggregation, we integrate the prediction distributions for each category in the whole image to construct the category correlation matrix, which is easier to be understood by the student network.

In summary, this paper makes the following contributions:

\begin{itemize}[topsep=0pt,itemsep=0pt]
    \item We propose a simple yet general and effective knowledge distillation framework called double similarity distillation for training accurate compact semantic image segmentation networks. Compared with other distillation methods, there are no extra parameters and only a minimal increase in FLOPs, which is more expansibility and generality.
    \item Pixel-wise similarity distillation module is proposed to capture detailed spatial dependencies through residual attention maps. Our module can be applied to multiple layers of the network simultaneously with low complexity, which is beneficial to learn the detailed spatial dependencies at different locations of the network.
    \item We design a category-wise similarity distillation module to extract global category correlation, which is calculated by constructing the category correlation matrix. Compared with pixel-based category correlation, the proposed module considers the characteristics of the segmentation task and avoids the interference of noise information.
\end{itemize}

To validate the effectiveness and generality of our method, we conduct extensive experiments on four challenging benchmark datasets, including Cityscapes~\cite{Cordts_2016_CVPR}, CamVid~\cite{brostow2008segmentation}, Pascal VOC 2012~\cite{everingham2010pascal}, and ADE20K~\cite{Zhou_2017_CVPR}. Comprehensive experiments and comparisons with state-of-the-arts demonstrate that our approach achieves very competitive performance.

\section{Related Work}
\label{sec:related}

Since our framework introduces the knowledge distillation strategy to improve the performance of the compact segmentation network, we briefly review some related works from three aspects: compact semantic image segmentation, knowledge distillation, and context information.

\subsection{Compact semantic image segmentation}

Compact semantic image segmentation is proposed to generate high-quality predictions with limited resources, which is more widely applied in edge devices and real time segmentation applications. One way to achieve the compact network is to introduce the model compression strategy, which can directly reduce the model size of cumbersome network. Such methods can be roughly categorized into weight pruning~\cite{Lebedev_2016_CVPR}, weight decomposition~\cite{zhang2015accelerating,NIPS2015_5787}, and weight quantization~\cite{Rastegari_xnor7,Jacob_2018_CVPR}. For example, INT8 Quantization~\cite{Jacob_2018_CVPR} proposed to rely only on 8bit integer arithmetic to approximate the floating-point computations. However, such methods may cause significant performance degradation. Moreover, some methods, like XNOR-Net~\cite{Rastegari_xnor7}, have low expansibility, because they are designed for special hardware. The second way is to utilize the compact classification backbone networks~\cite{Sandler_2018_CVPR,howard2017mobilenets,Ma_2018_ECCV,iandola2016squeezenet}, such as MobileNetV2~\cite{Sandler_2018_CVPR}, ShuffleNetV2~\cite{Ma_2018_ECCV}, and SqueezeNet~\cite{iandola2016squeezenet}. Most researches \cite{Zhao_2018_ECCV_icnet,romera2017erfnet,Li_2019_CVPR,7803544,paszke2016enet,Yu_2018_ECCV,Lin_2017_CVPR,Mehta_2018_ECCV,orsic2019defense} proposed to design a special compact semantic image segmentation network. For example, ENet~\cite{paszke2016enet} designed an efficient lightweight network, which used early downsampling operation to reduce the parameters. ICNet~\cite{Zhao_2018_ECCV_icnet} used multi-resolution branches cascading to enhance efficiency. BiSeNet~\cite{Yu_2018_ECCV} utilized two branches, one is a spatial path intended to learn spatial information and the other is context path aiming at obtaining the large receptive field. DFANet~\cite{Li_2019_CVPR} aggregated effective features through sub-network and sub-stage cascade. However, there is a certain margin in prediction accuracy between the performance of the compact semantic image segmentation networks and the state-of-the-art cumbersome networks. Beyond the above works, we exploit how to improve the performance of the compact semantic image segmentation networks based on the knowledge distillation strategy.

\subsection{Knowledge distillation}

The conventional knowledge distillation (KD) was originally proposed in~\cite{hinton2015distilling} to preserve the soft target of a complex ensemble of networks when adopting a compact network for more efficient deployment. Then, FitNets~\cite{adriana2015fitnets} proposed to transfer the intermediate representations. After FitNets, several methods~\cite{Zagoruyko2017AT,kim2018paraphrasing,yim2017gift,shen2019meal,heo2019knowledge} tried to transform the features to reduce the inherent differences of model architecture between two networks. For example, AT~\cite{Zagoruyko2017AT} transferred the attention maps by reducing the dimension, FT~\cite{kim2018paraphrasing} introduced the auto-encoder network to enhance the quality of feature maps, MEAL~\cite{shen2019meal} utilized the generative adversarial network to discriminate the features. Recently, some works~\cite{Park_2019_CVPR,Tung_2019_ICCV,Peng_2019_ICCV} proposed to distill the relationship between samples and the correlation of feature distribution, which also improve the performance of the compact network.

Since the successful application of knowledge distillation in the image classification, some works explored this strategy for many other domains, such as face recognition~\cite{luo2016face}, object detection~\cite{Wang_2019_CVPR}, lane detection~\cite{hou2019learning}, degraded image segmentation~\cite{guo2019degraded} and continual learning~\cite{8107520}. Hou~\cite{hou2019learning} proposed self-attention distillation to improve the representation
learning of lane detection models without any additional supervision. DGN~\cite{guo2019degraded} introduced the teacher-student architecture and proposed the dense-Gram loss to improve the segmentation performance on degraded images. LWF~\cite{8107520} exploited the knowledge distillation strategy to solve the catastrophic forgetting problem on the old tasks when learning new tasks. More recently, a number of research works have made use of knowledge distillation strategy to improve the compact semantic image segmentation networks~\cite{xie2018improving,He_2019_CVPR,Liu_2019_CVPR,liu2020structured,shan2019distilling}. MDE~\cite{xie2018improving} first proposed to match the probability representations and the local relationships between the student networks and the teacher networks. However, directly adopting these distillation methods designed for the task of image classification can only achieve sub-optimal improvements. KA~\cite{He_2019_CVPR}, SKD~\cite{Liu_2019_CVPR,liu2020structured}, and PFS~\cite{shan2019distilling} proposed to capture the spatial dependencies through Affinity module, while a large number of matrix multiplication operations limit its expansion and performance. To overcome this limitation, KA~\cite{He_2019_CVPR} utilized the auto-encoder network to reconstruct the representations. SKD~\cite{Liu_2019_CVPR,liu2020structured} learned soft targets and introduced the generative adversarial network to distill holistic knowledge. PFS~\cite{shan2019distilling} proposed to add the same distillation module on the teacher network to reduce the knowledge gap. However, these approaches introduce extra parameters and increase the training phases. Different from these methods, we propose a pixel-wise similarity distillation module through the subtraction operation to capture the spatial dependencies and category-wise similarity distillation module tailored for semantic image segmentation to extract the global category correlation. Note that our proposed method is still effective in sub-tasks of semantic image segmentation, such as lane detection~\cite{hou2019learning} and degraded image segmentation~\cite{guo2019degraded}.

\subsection{Context information}

\begin{figure*}[htp]
	\centering
	\includegraphics[width=.8\linewidth]{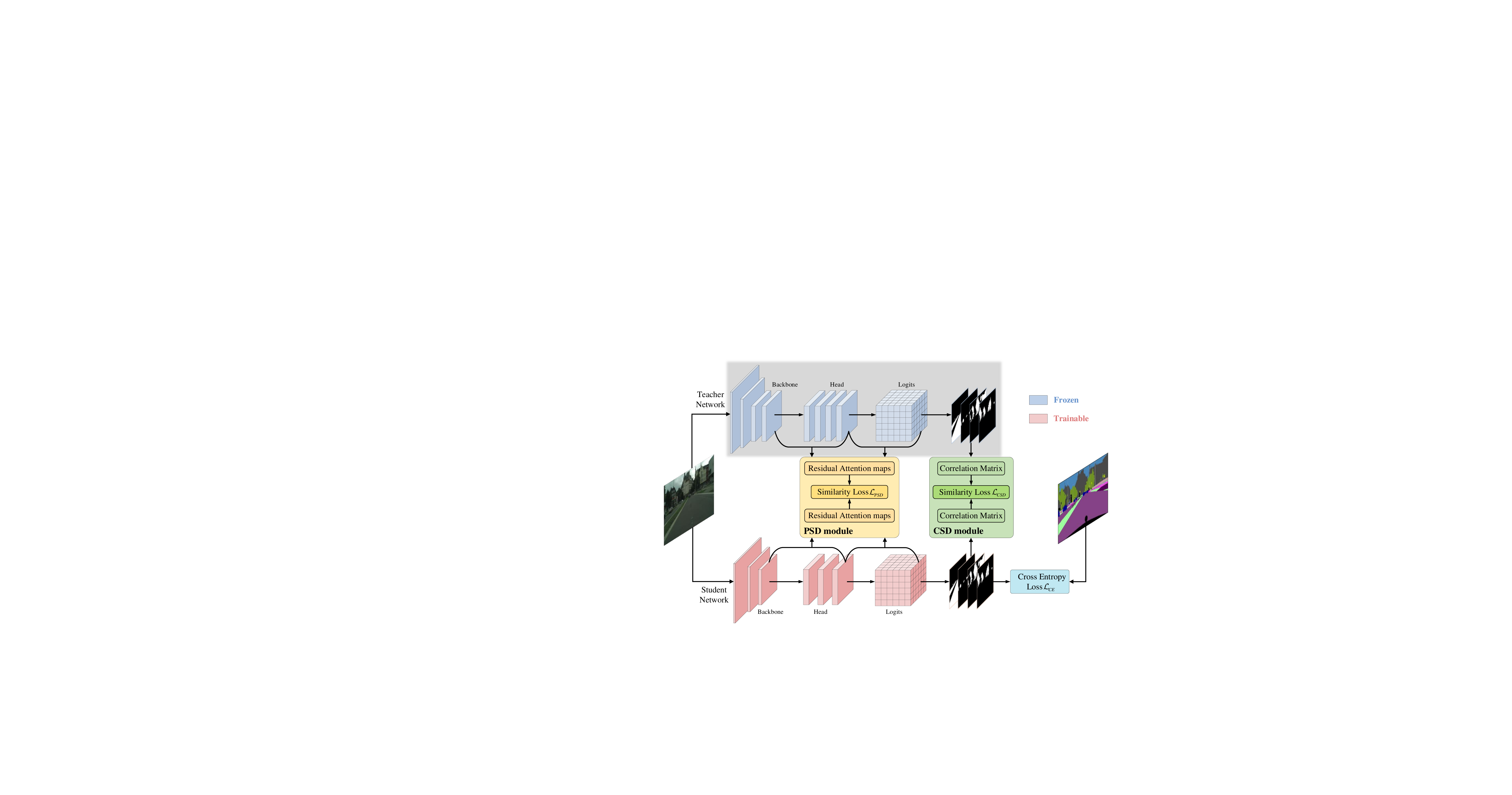}
	\caption{The pipeline of DSD framework. During the training process, teacher network is fixed and the student network is updated by the ground-truth labels and the knowledge transferred from the teacher network. The proposed pixel-wise similarity distillation module captures detailed spatial dependencies across the multiple layers of the network. The category-wise similarity distillation module strengthens the global category correlation. Best viewed in color.}
	\label{fig:framework}
\end{figure*}

Context information can capture long-range spatial dependencies and has recently been widely applied in semantic image segmentation. The direct way to get context information is to increase the receptive field of the network. Therefore, the dilated convolution~\cite{ding2018context,yu2015multi} and the pooling operation are widely adopted. For example, DeepLabv2~\cite{chen2017deeplab} proposed an atrous spatial pyramid pooling (ASPP), which consists of the dilated convolution with various dilation rates and the global pooling operation. PSPNet~\cite{zhao2017pyramid} proposed a different-region based pyramid pooling module to extract the context information. CCL~\cite{ding2018context} proposed a context contrasted local model to collect context contrasted local information. SVCNet~\cite{ding2019semantic} proposed a paired convolution to infer shape-variant context. More recently, based on the idea of pixel similarity, spatial dependence is proposed to achieve the start-of-the-art performance. For instance, DANet~\cite{Fu_2019_CVPR} proposed to learn semantic interdependencies in spatial and channel dimensions, respectively. BFP~\cite{ding2019boundary} proposed to harvest and propagate the local features to improve the similarity of the semantically homogenous pixel region and keep the discriminative power of the different pixel regions. Since spatial dependencies play an important role in semantic image segmentation, the previous distillation methods~\cite{He_2019_CVPR,Liu_2019_CVPR,liu2020structured,shan2019distilling} proposed to transfer such knowledge from the teacher network to the student network. However, a large number of matrix multiplication are introduced, which increases the complexity and limits further improvements in performance. Different from previous methods, we consider learning detailed spatial dependencies with low complexity.

\section{The Proposed Framework}
\label{sec:Methods}

In this section, we describe our contribution double similarity distillation framework. Section \ref{overview} introduces the overall framework. Section \ref{psd} and Section \ref{csd} introduce in detail the two distillation modules. Finally, Section \ref{opt} illustrates the training procedure of the proposed framework.

\subsection{Overview}
\label{overview}

The purpose of our framework is quite straightforward: Improving the classification accuracy of the student network through transferring knowledge from the teacher network. Given two deep convolution neural networks, a well-performed teacher network with parameters $\theta_t$ as $T$ and a new student network with parameters $\theta_s$ as $S$. The dataset is $\mathcal{D} = (\mathcal{X}, \mathcal{Y})$, where $ \mathcal{X}=\{x_1,x_2,...,x_n\}$ means the input images and $ \mathcal{Y}=\{y_1,y_2,...,y_n\}$ means the corresponding ground truth labels, $n$ denotes the number of samples in the dataset. Considering the training of the network from a probabilistic perspective, optimizing the parameters of student network $S$ given the dataset $\mathcal{D}$ is equivalent to maximizing the posterior probability $p(\theta_s|\mathcal{D})$. According to the Bayes’ rule, we can convert it to calculate the conditional probability $p(\mathcal{D}|\theta_s)$ and the prior probability of the parameters $p(\theta_s)$:
\begin{equation}
\log p(\theta_s|\mathcal{D}) = \log p(\mathcal{D}|\theta_s) + \log p(\theta_s) - \log p(\mathcal{D})
\label{eq:baye}
\end{equation}
where $p(\mathcal{D})$ represents the probability of the dataset, which is a constant independent of parameters. Therefore, the log probability of the data given the parameters $\log p(\mathcal{D}|\theta_s)$ can be regarded as the negative of the loss function $-\mathcal{L}_{\theta_s}$. When the teacher network is introduced to train the student network, the optimization of parameters can be redefined as follows:
\begin{equation}
\begin{split}
\log p(\theta_s|\mathcal{D},\theta_t) &= \log p(\mathcal{D},\theta_s|\theta_t) + \log p(\theta_t) - \log p(\mathcal{D},\theta_t)\\
&=\log p(\mathcal{D}|\theta_s,\theta_t) + \log p(\theta_s|\theta_t) - \log p(\mathcal{D}|\theta_t)
\end{split}
\end{equation}
Note that the parameters of the teacher network have been optimized and fixed. The log probability of the data given the parameters of the teacher network $-\log p(\mathcal{D}|\theta_t)$ ($\mathcal{L}_{\theta_t}$) should be an extremely small constant. Therefore, the log probability $\log p(\theta_s|\theta_t)$ corresponds to the distillation loss function $\mathcal{L}_{distill}$. Compared with Eq.~\ref{eq:baye}, transferring comprehensive and powerful knowledge from the teacher network is essential to improve the performance of the student network.

Based on the above consideration, we design a system that combines pixel-wise similarity and category-wise similarity to improve the performance of the compact semantic image segmentation network. The overall framework of the proposed double similarity distillation, called DSD, is shown in Fig.~\ref{fig:framework}. The semantic image segmentation networks are FCN-like architecture~\cite{long2015fully}, which consists of the backbone network, head network, and logits layer in sequence. The backbone network takes one image as the input to extract its feature maps. Then, the head network can aggregate higher-level information based on the output of the backbone network. Finally, the logits layer is applied to generate the class probability values. Generally, the teacher network is frozen, which applies complex and heavy models to produce higher prediction accuracy. The student network is trained with the ground truth labels and the knowledge transferred through the proposed PSD module and CSD module.

\subsection{Pixel-wise Similarity Distillation Module}
\label{psd}

The spatial dependencies play a powerful role in semantic image segmentation. The existing distillation methods~\cite{He_2019_CVPR,Liu_2019_CVPR,liu2020structured,shan2019distilling} utilized the Affinity module to capture the information. Given the feature map $A \in R^{N \times H \times W}$, where $N$, $H$ and $W$ denote the number of channel, height and width, respectively. The spatial dependence can be defined as follows:
\begin{equation}
s_i =\Psi\left(\frac{1}{Z}\sum\nolimits_{j=1}^{Z}f(a_i,a_j)\right),Z=H \times W
\end{equation}
where $Z$ equals to $H \times W$, $a_i,a_j$ denote the features corresponding to pixel $i$ and pixel $j$, respectively. $\Psi(\cdot)$ means the activation function, $f(a_i,a_j)$ represents the pairwise function that calculates the affinity between pixel feature vector $a_i$ and $a_j$. In KA~\cite{He_2019_CVPR} and SKD~\cite{Liu_2019_CVPR,liu2020structured}, the pairwise function is implemented as follows:
\begin{equation}
f(a_i,a_j) = \mathcal{N}(a_i)^T\mathcal{N}(a_j)
\end{equation}
here $\mathcal{N}(v)=v/\|v\|_2$ denotes the $l_2$-normalized feature vector. Therefore, ignoring the activation function and normalization operation, the Affinity module in KA~\cite{He_2019_CVPR} and SKD~\cite{Liu_2019_CVPR,liu2020structured} has a computational cost of:
\begin{equation}
(2N-1) \times H \times W \times H \times W 
\end{equation}
the high complexity comes from a large number of matrix multiplication. Therefore, it is hard to capture more detailed spatial dependencies on multiple layers of the network, which limits its application.

\begin{figure}[t]
	\centering
	\subfigcapskip=5pt
	\includegraphics[width=.9\linewidth]{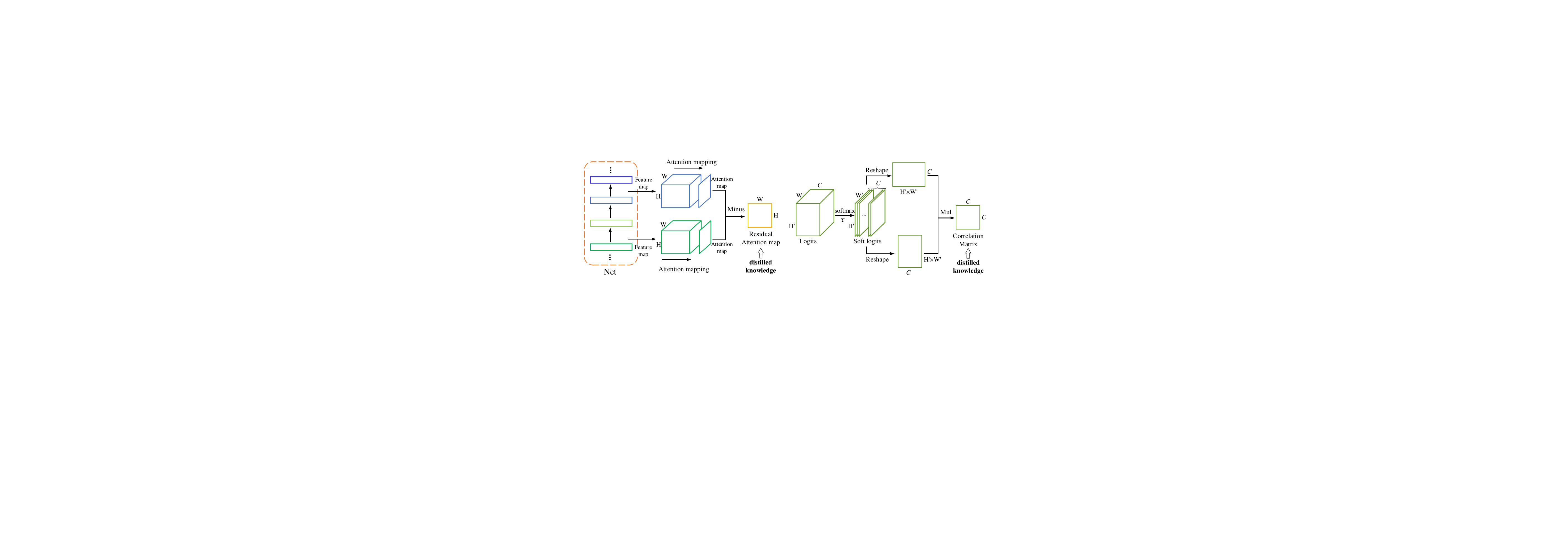}
	\caption{Schematic diagram of the pixel-wise similarity distillation module. The residual attention map is calculated by the subtraction between any two attention maps, and the attention map is obtained by attention mapping operation on feature maps.}
	\label{fig:ps_detail}
\end{figure}

Inspired by the residual learning~\cite{he2016deep} and attention transfer~\cite{Zagoruyko2017AT}, we propose the pixel-wise similarity distillation module to construct the residual attention (RA) map. As shown in Fig.~\ref{fig:ps_detail}, the residual attention map can be calculated through subtraction between attention maps, which are obtained by attention mapping operation on feature maps. The responses of pixels belonging to the same category on the residual attention map should be similar. Therefore, the residual attention map encodes spatial structural knowledge, which is helpful for student network to learn the spatial dependencies.

The PSD module can be applied across multiple convolutional layers. Let $\mathcal{I} = \{A^1, A^2, ..., A^K\}$ represents a set of K selected feature maps. First, we can get the attention maps through the attention mapping function:
\begin{equation}
\mathcal{F}(A): R^{N \times H \times W} \rightarrow R^{H \times W}, A \in \mathcal{I}
\end{equation}
Specifically, the attention mapping function can be divided into two categories~\cite{Zagoruyko2017AT}, i.e. maximum operation $\mathcal{F}_{max}^p(A) = \max_{i=1,N}|a_i|^p$ and summation operation $\mathcal{F}_{sum}^p(A) = \sum_{i=1}^{N}|a_i|^p$. Here $p (p \geqslant 1)$ denotes the power factor. Following the previous works~\cite{Zagoruyko2017AT,hou2019learning}, we choose the summation operation and set $p$ as 2 to obtain the mapping function $\mathcal{F}_{sum}^2(\cdot)$. Then, the residual attention map can be defined as follows:
\begin{equation}
RA^{mn}
= \mathcal{N}(\mathcal{F}_{sum}^2(A^{m})) - \mathcal{N}(\mathcal{F}_{sum}^2(A^n)), m>n
\label{RA}
\end{equation}
Here, $A^m$ and $A^n$ are adjacent feature maps in the set $\mathcal{I}$. The bilinear operation will be added if the size of feature maps in different spatial dimensions. Similarly, we can get the computational cost of the PSD module:
\begin{equation}
(2K \times N -1) \times H \times W
\end{equation}

Compared with the computational cost of Affinity module, we get a reduction in the computation of:
\begin{equation}
\frac{(2K \times N -1) \times H \times W}{(2N-1) \times H \times W \times H \times W } \approx \frac{K}{H \times W} = \frac{K}{Z}
\end{equation}
Considering that $K \ll Z $, the PSD module is thousand times less computation than Affinity module, which allows our module to capture detailed spatial dependencies on multiple layers.

We adopt $\mathcal{L}_2$ loss to calculate the pixel-wise similarity distillation loss between the teacher and the student networks, which is formulated as follows:
\begin{equation}
\begin{split}
\mathcal{L}_{PSD} &= \sum_{m,n \in \mathcal{I}}\mathcal{L}_2 \left(\mathcal{N}(RA_S^{mn}), \mathcal{N}(RA_T^{mn})\right)\\
&= \frac{1}{(K-1)Z}\sum_{m,n \in \mathcal{I}} \left\|\frac{RA_S^{mn}}{\|RA_S^{mn}\|_2} - \frac{RA_T^{mn}}{\|RA_T^{mn}\|_2}\right\|^2_2
\end{split}
\label{eq:psd}
\end{equation}
where $RA_T$ represents the residual attention map of the teacher network and $RA_S$ represents the corresponding residual attention map of the student network. Note that the complexity of all methods is only calculated before the $L_2$ loss, because the $L_2$ distance loss function is also applied to our comparison methods.

\begin{figure}[t]
	\centering
	\subfigcapskip=5pt
	\subfigure{\begin{minipage}{.31\linewidth}
			\centering
			\includegraphics[width=\linewidth,height=.65\linewidth]{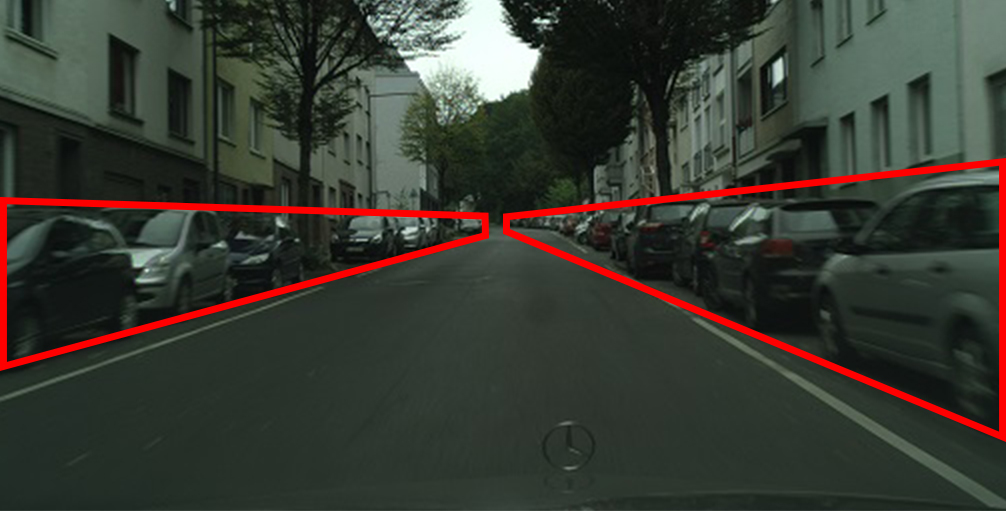}
	\end{minipage}}
	\hskip -1pt
	\subfigure{\begin{minipage}{.31\linewidth}
			\centering
			\includegraphics[width=\linewidth,height=.65\linewidth]{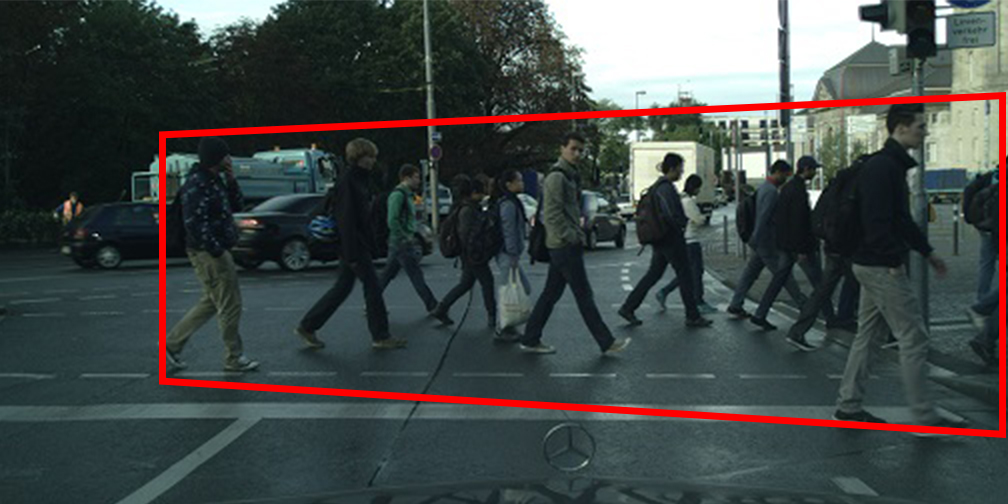}
	\end{minipage}}
	\hskip -1pt
	\subfigure{\begin{minipage}{.31\linewidth}
			\centering
			\includegraphics[width=\linewidth,height=.65\linewidth]{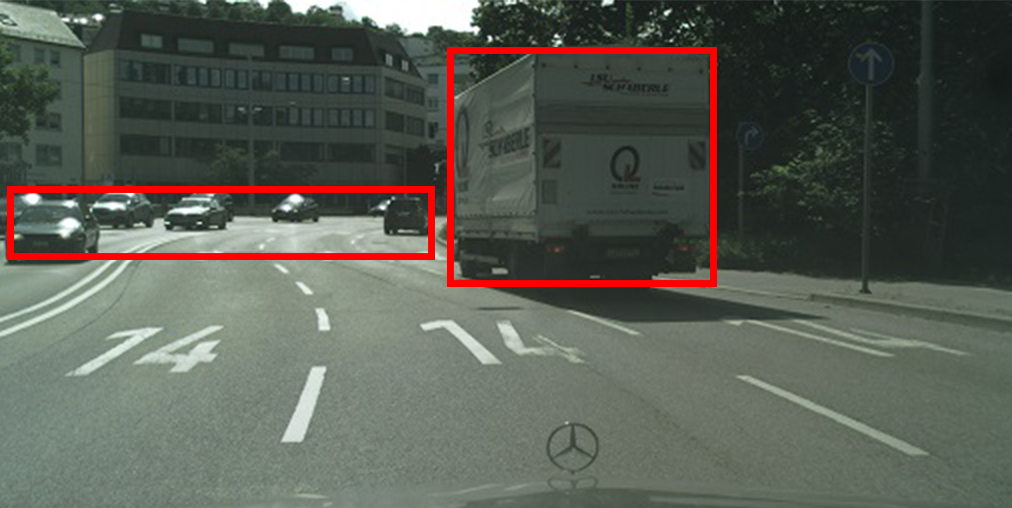}
	\end{minipage}}
	\vskip -7pt
	\subfigure{\begin{minipage}{.31\linewidth}
			\centering
			\includegraphics[width=\linewidth,height=.65\linewidth]{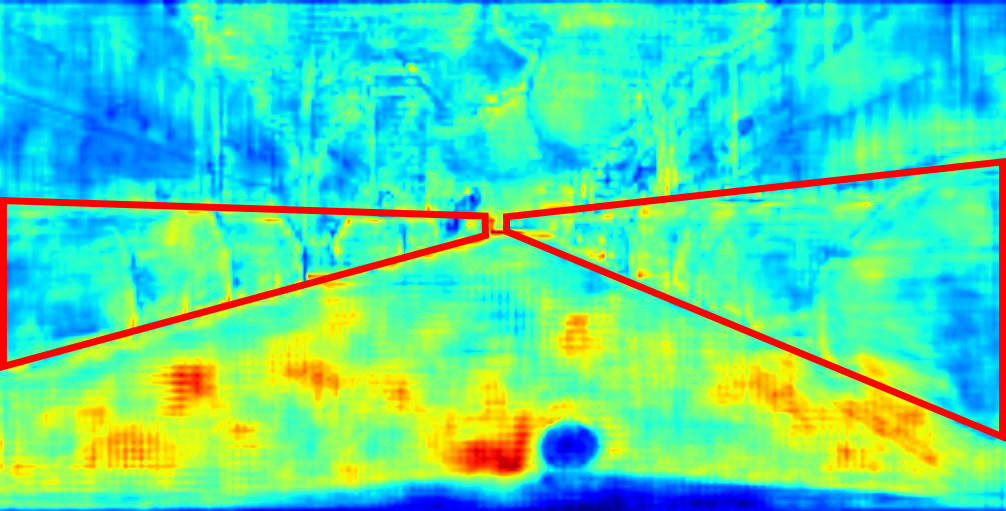}
	\end{minipage}}
	\hskip -1pt
	\subfigure{\begin{minipage}{.31\linewidth}
			\centering
			\includegraphics[width=\linewidth,height=.65\linewidth]{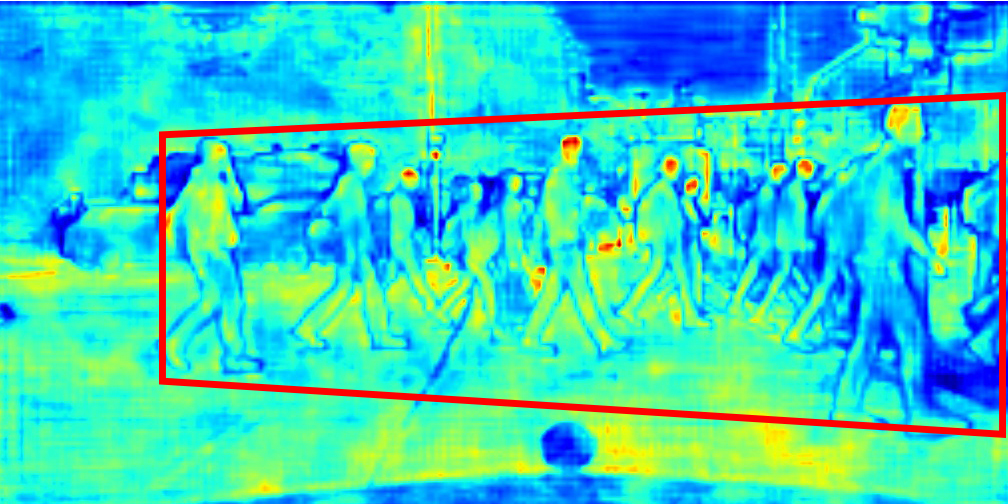}
	\end{minipage}}
	\hskip -1pt
	\subfigure{\begin{minipage}{.31\linewidth}
			\centering
			\includegraphics[width=\linewidth,height=.65\linewidth]{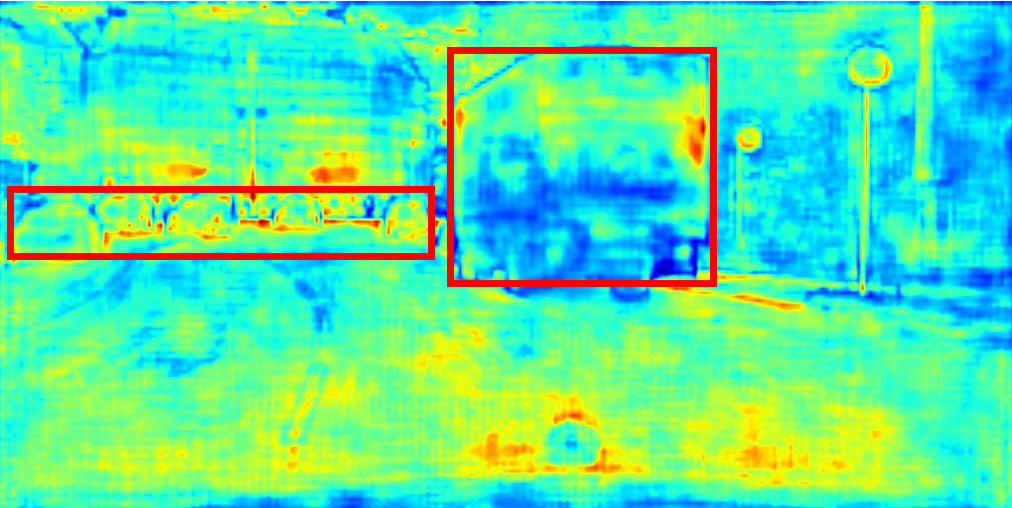}
	\end{minipage}}
	\vskip -7pt
	\subfigure{\begin{minipage}{.31\linewidth}
			\centering
			\includegraphics[width=\linewidth,height=.65\linewidth]{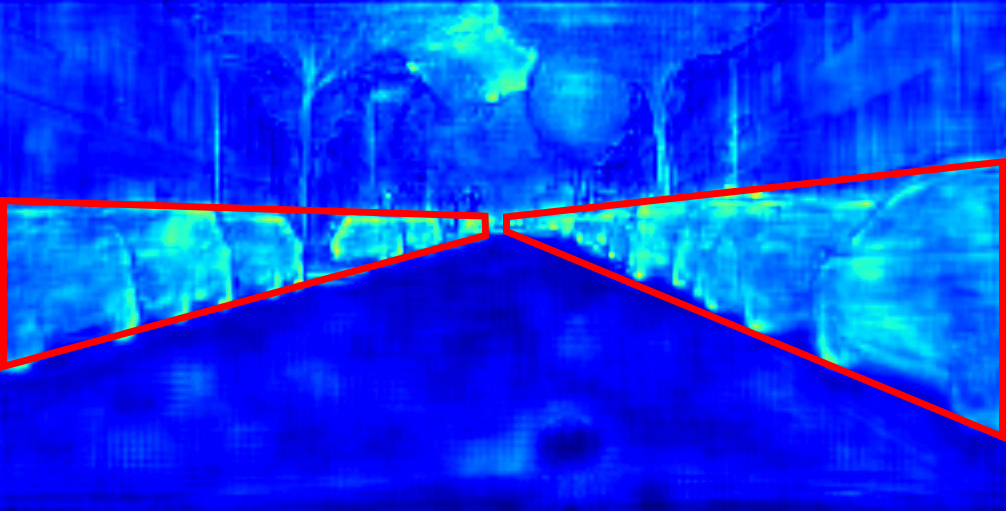}
	\end{minipage}}
	\hskip -1pt
	\subfigure{\begin{minipage}{.31\linewidth}
			\centering
			\includegraphics[width=\linewidth,height=.65\linewidth]{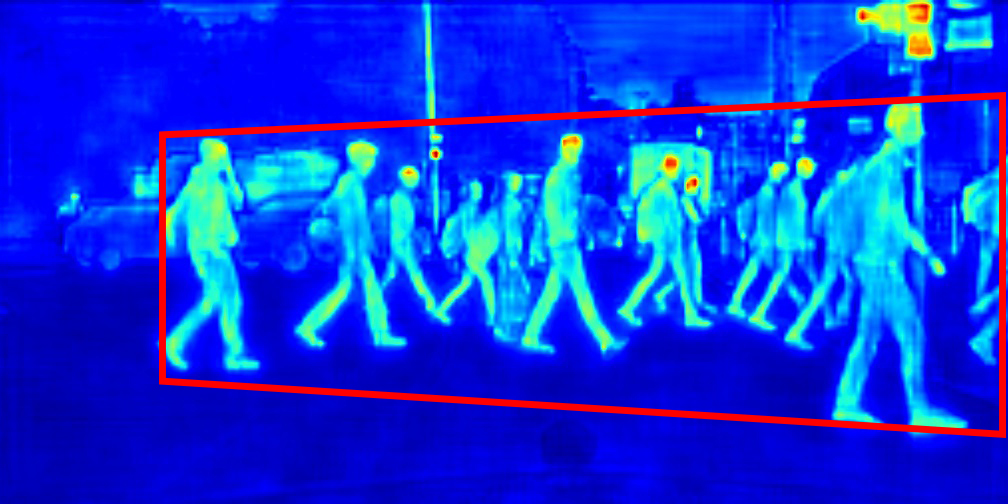}
	\end{minipage}}
	\hskip -1pt
	\subfigure{\begin{minipage}{.31\linewidth}
			\centering
			\includegraphics[width=\linewidth,height=.65\linewidth]{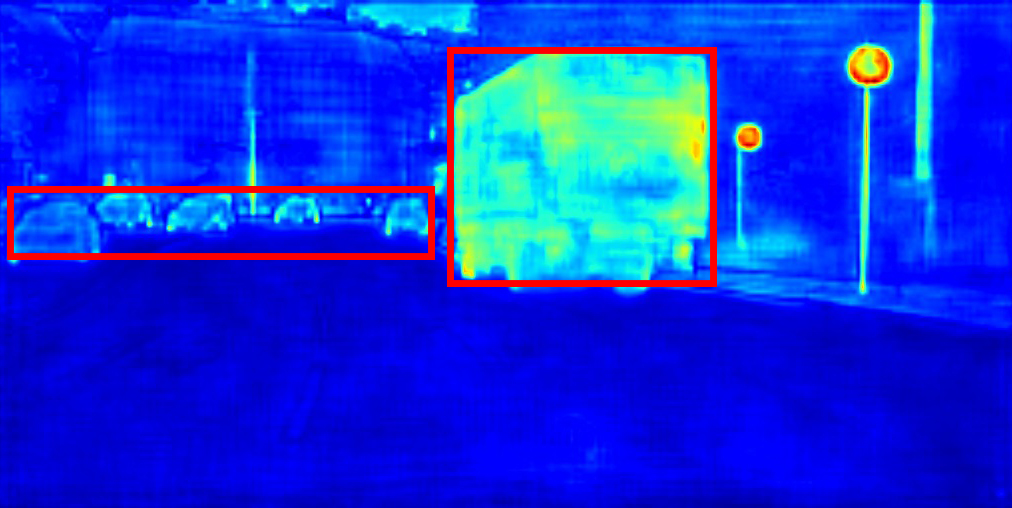}
	\end{minipage}}
	\vskip -7pt
	\subfigure{\begin{minipage}{.31\linewidth}
			\centering
			\includegraphics[width=\linewidth,height=.65\linewidth]{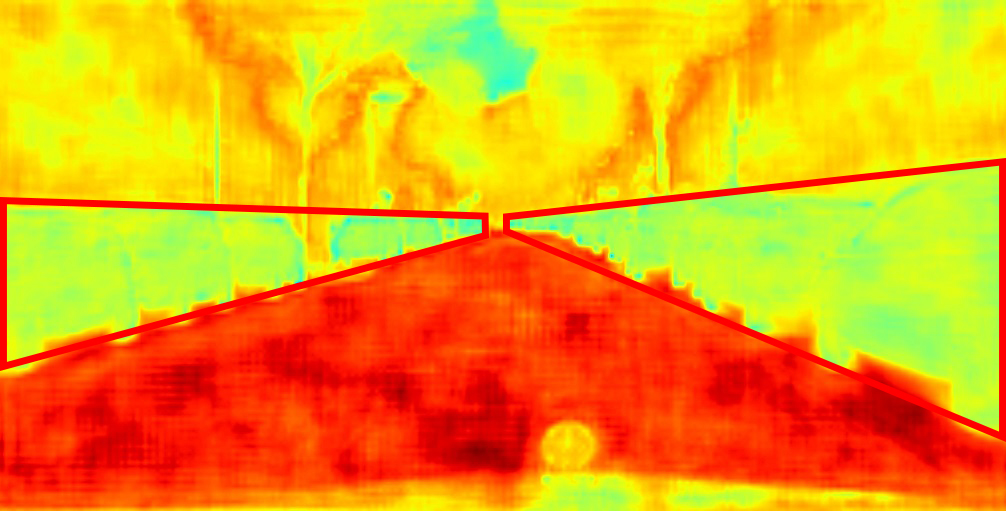}
	\end{minipage}}
	\hskip -1pt
	\subfigure{\begin{minipage}{.31\linewidth}
			\centering
			\includegraphics[width=\linewidth,height=.65\linewidth]{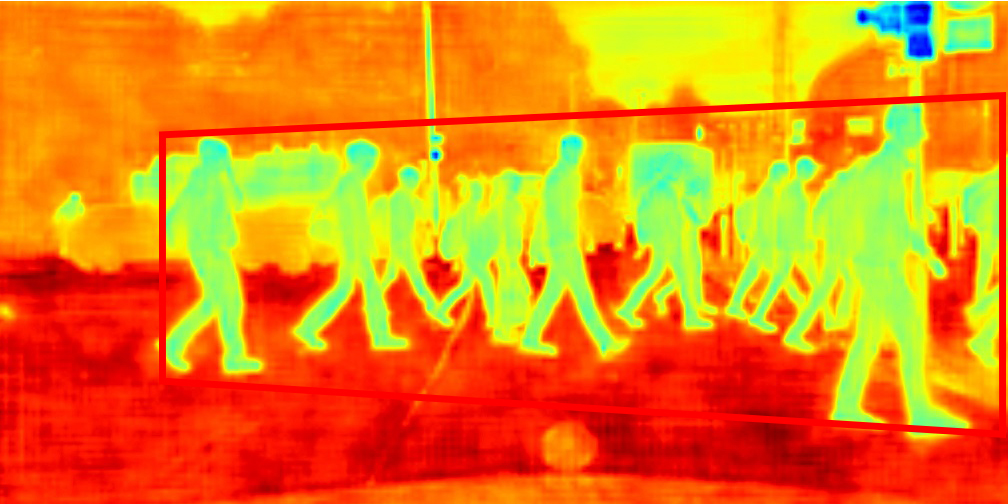}
	\end{minipage}}
	\hskip -1pt
	\subfigure{\begin{minipage}{.31\linewidth}
			\centering
			\includegraphics[width=\linewidth,height=.65\linewidth]{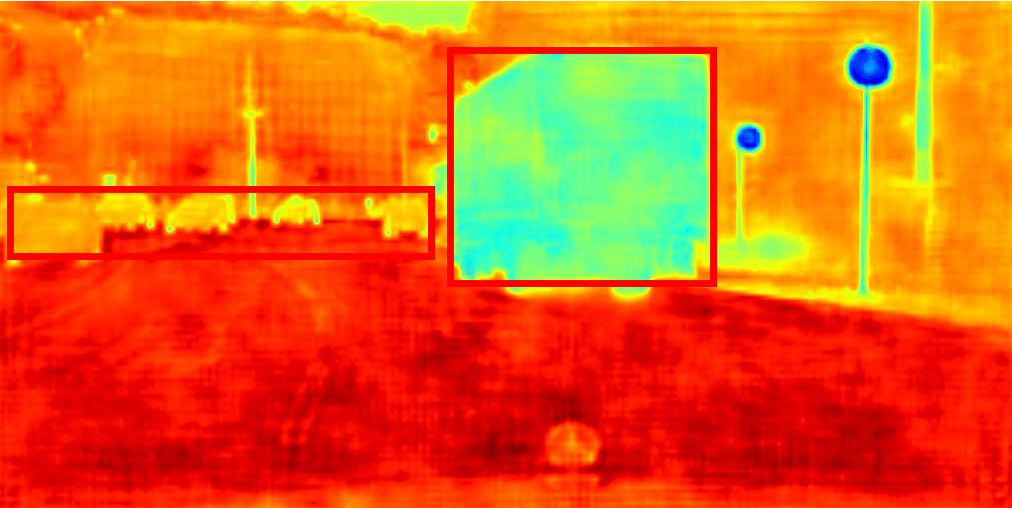}
	\end{minipage}}
	\caption{Visualization of the attention maps and the corresponding residual attention maps. The images in the first row represent the input images; the ones in the second row represent the attention maps generated by the output of the logits layer; the ones in the third row represent the attention maps generated by the output of the head network; the images in the last row represent the corresponding residual attention maps. Best viewed in color.}
	\label{fig:fa}
\end{figure}

To visualize the effect of the PSD module, some examples of attention maps and the corresponding residual attention maps are presented in Fig.~\ref{fig:fa}. It can be seen that the PSD module successfully captures spatial dependencies. Note that the PSD module only requires less computational complexity. Meanwhile, Fig.~\ref{fig:fa} also shows why we do not directly use attention transfer~\cite{Zagoruyko2017AT} for semantic image segmentation task. The attention maps can only generate local discrimination for objects. For example, the high response of pixels in the head of the person and the high response of pixels in the outline of the car. The attention transfer encourages the student network to mimic the attention maps of the teacher network and proves its effectiveness in image classification tasks. This is because the classification networks are inclined to identify patterns from the most discriminative parts for recognition. However, the semantic image segmentation needs localize integral regions of the objects densely within the whole image to generate accuracy prediction results.

\subsection{Category-wise Similarity Distillation Module}
\label{csd}

If the segmentation task is taken as a collection of separate pixel classification tasks, and KD~\cite{hinton2015distilling} is used to distill the category correlation knowledge, only sub-optimal improvements can be obtained, such as the experiments in Section~\ref{exp}. For one image, pixels of the same category may produce very different prediction distributions and obtain contradictory category correlation. Therefore, we propose to integrate the distributions for each pixel of the same category and strengthen the global category correlation based on the perspective of global aggregation.

\begin{figure}[t]
	\centering
	\subfigcapskip=5pt
	\includegraphics[width=.9\linewidth]{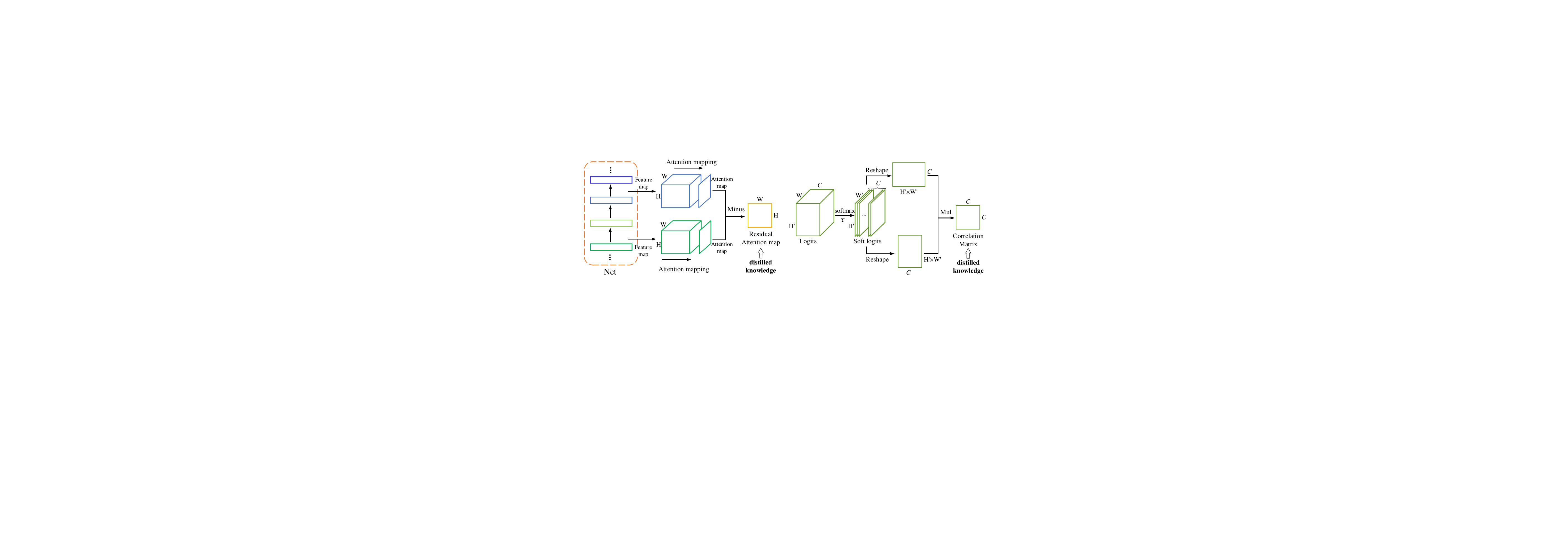}
	\caption{Schematic diagram of the category-wise similarity distillation module. The CSD module is applied to the soft logits layer and utilizes the self-attention mechanism to construct the correlation matrix.}
	\label{fig:cs_detail}
\end{figure}

At the logits layer, the number of channels equals the number of categories, the channels and categories are corresponding one by one. Therefore, the values in one specific channel represent the corresponding class probabilities of all pixels. As shown in Fig.~\ref{fig:cs_detail}, the self-attention mechanism~\cite{Wang_2018_CVPR} between any two channels at the soft logits layer can be applied to construct the correlation matrix, that is, the global category-wise similarity knowledge, which can help students learn the global category correlation.

Explain the CSD module more formally. Let's denote the probability values of the logits layer by $z \in R^{C \times H'W'}$ with $C$ categories and spatial dimensions ${H' \times W'}$. Then, we can get the soft logits $q = softmax({z}/{\tau})$, where $\tau$ is the temperature to soften the output. For better illustration, the soft logits of each category is expressed as $q^{k\cdot} \in R^{H'W'}$. The correlation matrix $CM \in R^{C \times C}$ can be defined as follows:
\begin{equation}
CM(q)_{ij}= \mathcal{N}(q^{i\cdot}) \cdot \mathcal{N}(q^{j\cdot})
\end{equation}
where $i$ and $j$ are the index for the category. The computational cost of CSD module:
\begin{equation}
(2H' \times W' -1) \times C \times C
\end{equation}

The $\mathcal{L}_2$ loss is introduced to calculate the category-wise similarity distillation loss between the teacher and student networks, which is defined as follows:
\begin{equation}
\begin{split}
\mathcal{L}_{CSD} &= \mathcal{L}_2 (CM_S(q), CM_T(q))\\
&=\frac{1}{C^2}\Big\|CM_S(q) - CM_T(q)\Big\|^2_2
\end{split}
\label{eq:csd}
\end{equation}
where $CM_T$ represents the correlation matrix of teacher network and $CM_S$ represents the corresponding correlation matrix of the student network.

\subsection{Optimization}
\label{opt}

\begin{algorithm}[t]
	\caption{Double Similarity Distillation (DSD).}
	\label{code:dsd}
	\begin{algorithmic}[1]
		\REQUIRE ~~\\
		$T$, $\theta_t$ are the per-trained teacher network and the corresponding parameters;\\
		$\mathcal{I}$ is the set of selected feature maps, the size is $K$;\\
		$\alpha$, $\beta$ are the weight parameters of different objects;\\
		$\mathcal{D}=\{\mathcal{X},\mathcal{Y}\}$ is the training data, $\mathcal{X},\mathcal{Y}$ mean the input images and corresponding ground truth labels;
		\ENSURE ~~\\
		$S$, $\theta_s$ are the student network and the corresponding parameters;
		\STATE Initialize the student network $S$ and load weights for the teacher network $T$.
		\FOR {each mini-batch $\mathcal{D}_b=\{\mathcal{X}_b,\mathcal{Y}_b\}$}
		\STATE Employ the teacher network $T$ on the mini-batch $\mathcal{D}_b$:
		\STATE \hspace{2cm} {$\{RA_T,CM_T\}\leftarrow T(\mathcal{X}_b, \theta_t)$};
		\STATE Employ the student network $S$ on the mini-batch $\mathcal{D}_b$:
		\STATE \hspace{1.8cm} {$\{RA_S,CM_S,\mathcal{Y}'_b\}\leftarrow S(\mathcal{X}_b,\theta_s)$};
		\STATE Calculate the pixel-wise similarity distillation loss $\mathcal{L}_{PSD}$ (Eq. \ref{eq:psd});
		\STATE Calculate the category-wise similarity distillation loss $\mathcal{L}_{CSD}$ (Eq. \ref{eq:csd});
		\STATE Calculate the cross-entropy loss $\mathcal{L}_{CE}$;
		\STATE \hspace{2.4cm} {$\mathcal{L}_{CE} = -\sum \mathcal{Y}_b \log \mathcal{Y}'_b$};
		\STATE Calculate the total loss $\mathcal{L}_{Total}$ (Eq. \ref{eq:total});
		\STATE Update the parameters $\theta_s$ of the student network $S$:
		\STATE \hspace{2.5cm} {$\theta_s = argmin_{\theta_s} \mathcal{L}_{total}$};
		\ENDFOR
		\RETURN  $S$, $\theta_s$.
	\end{algorithmic}
\end{algorithm}

The proposed double similarity distillation framework consists of the teacher network and the student network, which is designed to transfer the similarity knowledge in pixel and category dimensions, respectively. During the training process, the parameters of the teacher network will be frozen. The student network is trained with the task-specific loss $\mathcal{L}_{\theta_s}$ and the distillation loss $\mathcal{L}_{distill}$. Specifically, the former is the standard cross-entropy loss function $\mathcal{L}_{CE}$ and the latter is composed of two proposed distillation loss functions $\mathcal{L}_{PSD}$ and $\mathcal{L}_{CSD}$. The total loss is defined as follows:
\begin{equation}
\begin{split}
\mathcal{L}_{Total} &= \mathcal{L}_{\theta_s} + \mathcal{L}_{distill} \\
&= \mathcal{L}_{CE} + \alpha \mathcal{L}_{PSD} + \beta \mathcal{L}_{CSD}
\end{split}
\label{eq:total}
\end{equation}
where $\alpha$ and $\beta$ are loss weights to make these loss values ranges comparable. The pipeline of DSD framework can be seen in Algorithm~\ref{code:dsd}. There are no extra parameters and negligible FLOPs, which makes the proposed method more expansibility and generality. Besides, the proposed method has no constraints on the semantic image segmentation network and can be easily implemented in an end-to-end manner.

\section{Experiments}
\label{exp}

In this section, we first provide a concise description of the experimental details, such as datasets, experiment setup, and evaluation metrics. Subsequently, we analyze and evaluate the effectiveness of our method. Finally, we show results on four benchmarks and compare them with state-of-the-art methods.

\subsection{Dataset}

We evaluate our proposed framework on four standard semantic segmentation datasets: Cityscapes~\cite{Cordts_2016_CVPR}, CamVid~\cite{brostow2008segmentation}, Pascal VOC 2012~\cite{everingham2010pascal}, and ADE20K~\cite{Zhou_2017_CVPR}.

\subsubsection{Cityscapes}

The Cityscapes~\cite{Cordts_2016_CVPR} is a large-scale urban street scenes dataset, which contains 19 classes that are used for evaluation. It contains 5000 high-quality pixel-level annotations images with a resolution of $1024 \times 2048$, and these finely annotated images are divided into 2975 training images, 500 validation images, and 1525 test images.

\subsubsection{CamVid}

The CamVid~\cite{brostow2008segmentation} is a small scene understanding dataset, which contains 367 training images and 233 testing images. The dataset contains 11 different categories and one ignore label for unlabeled pixels.

\subsubsection{Pascal VOC 2012}

The Pascal VOC 2012~\cite{everingham2010pascal} is a popular object-centric segmentation dataset with 20 object classes and one background class. It contains 1464 training images, 1449 validation images, and 1456 test images. Following the common setting~\cite{zhao2017pyramid,chen2018encoder,Zhao_2018_ECCV}, we use the additionally annotated images resulting in 10582 training images.

\subsubsection{ADE20K}

We further verify the effectiveness of our method on the ADE20K dataset~\cite{Zhou_2017_CVPR}. The ADE20K is a complex and challenging scene parsing dataset, which contains 150 semantic categories. It includes 20K/2K/3K images for training, validation, and testing, respectively.

\subsection{Implementation Details}

\subsubsection{Teacher Network}

PSPNet~\cite{zhao2017pyramid} is one of the state-of-the-art semantic image segmentation frameworks, which proposed Pyramid Pooling Module (PPM) to gather both local and global context information into the final feature representation through four different parallel pooling branches. Its effectiveness has been verified in several segmentation benchmarks. Therefore, we select PSPNet~\cite{zhao2017pyramid} with ResNet-101~\cite{he2016deep} as the teacher network. Specifically, the backbone and head network of the teacher are ResNet-101 and PPM, respectively. Besides, we employ the open-source trained models~\cite{semseg2019} of the author as the teacher models.

\subsubsection{Student Network}

There are no restrictions on student networks. Two popular compact networks are selected as the student models, one is ResNet-18~\cite{he2016deep} and the other is MobileNetV2~\cite{Sandler_2018_CVPR}. The former has the same residual structure as ResNet-101~\cite{he2016deep}, and the only difference is the depth of the network. The latter utilizes inverted residuals and linear bottlenecks, which is completely different from the teacher network. Therefore, the two student networks are used to verify the effectiveness under the condition of similar network structures and different network structures, respectively.

\subsubsection{Training setup}

Our network is trained end-to-end with stochastic gradient descent (SGD) with batch size 16 (at most case), momentum 0.9, and weight decay 0.0001. The pre-trained models are trained on the ImageNet~\cite{russakovsky2015imagenet} and the output stride is set to 1/8. Following previous works~\cite{zhao2017pyramid,Zhao_2018_ECCV,Yu_2018_ECCV}, we use the poly strategy in which the current learning rate is multiplied by $(1 - \frac{iter}{max\_iter})^{power}$ each iteration with power 0.9. The base learning rate is set to 0.01. Following the previous works~\cite{Liu_2019_CVPR,liu2020structured,He_2019_CVPR}, the weight balance parameters $\alpha$ and $\beta$ are set to $10^3$ and 10 to maintain a balance with the cross-entropy loss function. The iteration number is set to 40K for Cityscapes, 10K for CamVid, 60K for Pascal VOC 2012, and 125K for ADE20K. The input size is set to $513 \times 513$ for Cityscapes, $480 \times 360$ for CamVid, and $473 \times 473$ for both Pascal VOC 2012 and ADE20K during training and inference. For data augmentation, the random horizontal flip and random resize between 0.5 and 2 are adopted. Then, all images are resized to have the maximum extent of the long side of the input size and padded with mean to the input size. For inference, we verify the performance on a single scale and original inputs. Besides, only the conventional cross-entropy loss is applied in the experiments, the class probability weighting, hard sample mining strategy, and deep supervision~\cite{zhao2017pyramid} are excluded unless otherwise specified. Our method is implemented by Pytorch~\cite{paszke2017automatic} framework. All networks are trained on a single NVIDIA Tesla V100 GPU with 32GB memory.

\subsubsection{Evaluation metrics}

The mean Intersection over Union (mIoU) is introduced to evaluate the effectiveness of the proposed method on four benchmark datasets, which is defined as follows:
\begin{equation}
\text{mIoU} = \frac{1}{C} \sum_{i=1}^{C} \frac{p_{ii}}{\sum_{j=1}^{C} p_{ij}+\sum_{j=1}^{C}p_{ji}-p_{ii}}
\end{equation}
where $C$ is the number of categories, $p_{ij}$ is the number of pixels of class $i$ predicted as belonging to class $j$. The numbers of true positives $p_{ii}$ represents the intersection. The sum of true positives $p_{ii}$, false negatives $p_{ji}$, and false positives $p_{ij}$ represents the union. The Intersection over Union (IoU) is calculated as the ratio of intersection and union between the ground truth and the prediction results for each class. The mIoU is calculated on a per-class basis and then averaged. We also report the pixel-wise accuracy (Pixel Acc.) to investigate the accuracy on ADE20K dataset, which is formulated as follows:
\begin{equation}
\text{Pixel Acc.} = \frac{\sum_{i=1}^{C}p_{ii}}{\sum_{i=1}^{C}\sum_{j=1}^{C}p_{ij}}
\end{equation}
Here, the numerator represents the properly classified pixels and the denominator denotes the total number of pixels. Besides, the number of float-point operations (FLOPs) is applied to measure the computational complexity, and the model size is evaluated by the network parameters (Params).

\subsection{Ablation Studies}

In this section, we show different the effectiveness of our method with comprehensive ablation experiments on the Cityscapes validation set. The effectiveness of the proposed DSD framework is provided in Table~\ref{table:ablation1}. The Table~\ref{table:ablation2} and Table~\ref{table:ablation4} detail the effect of the PSD module. The table~\ref{table:ablation3} shows the effect of the CSD module.

\subsubsection{Effectiveness of double similarity distillation}

Two student networks are used to explore the effectiveness of the DSD framework. As shown in Table \ref{table:ablation1}, compared with the original student networks, the performance of the MobileNetV2 and ResNet-18 can be improved by 3.67 and 3.79 points, respectively. This implies that the proposed DSD framework successfully helps the compact segmentation network achieve a significant improvement. Specifically, The PSD module brings 1.90\% improvement for MobileNetV2 and 2.82\% improvement for ResNet-18, respectively. Note that although the architecture of student network MobileNetV2 is completely different from the architecture of the teacher network ResNet-101, the PSD module still works well. This means that PSD is well generalized. Besides, the CSD module can improve the mIoU score based on the PSD module, which can be complementary to the PSD module. This proves that the proposed DSD framework can distill comprehensive and complementary similarity knowledge. Therefore, the proposed method can be simple and effectively applied to all existing compact semantic image segmentation networks.

\begin{figure*}[htbp]
	\centering
	\subfigcapskip=5pt
	\subfigure{\begin{minipage}{.13\linewidth}
			\centering
			\includegraphics[width=\linewidth]{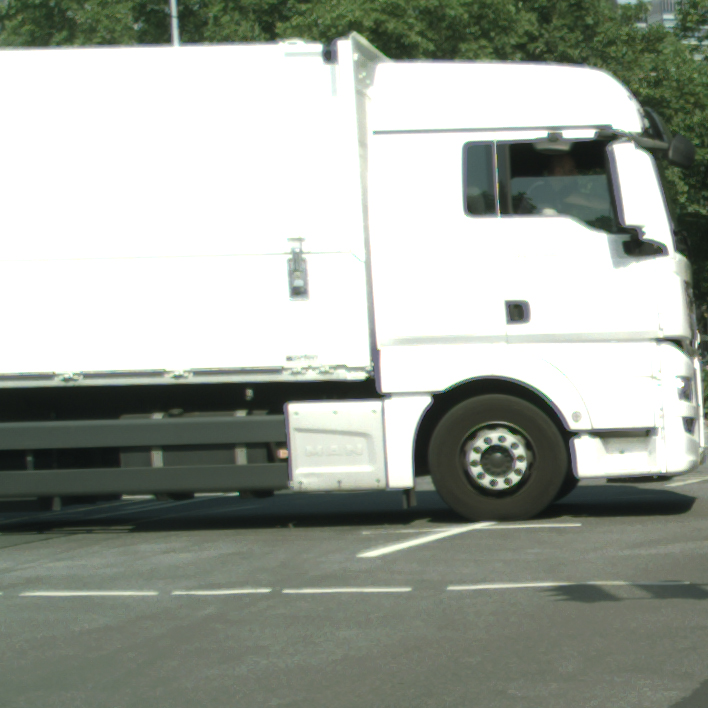}
	\end{minipage}}
	\hskip -1pt
	\subfigure{\begin{minipage}{.13\linewidth}
			\centering
			\includegraphics[width=\linewidth]{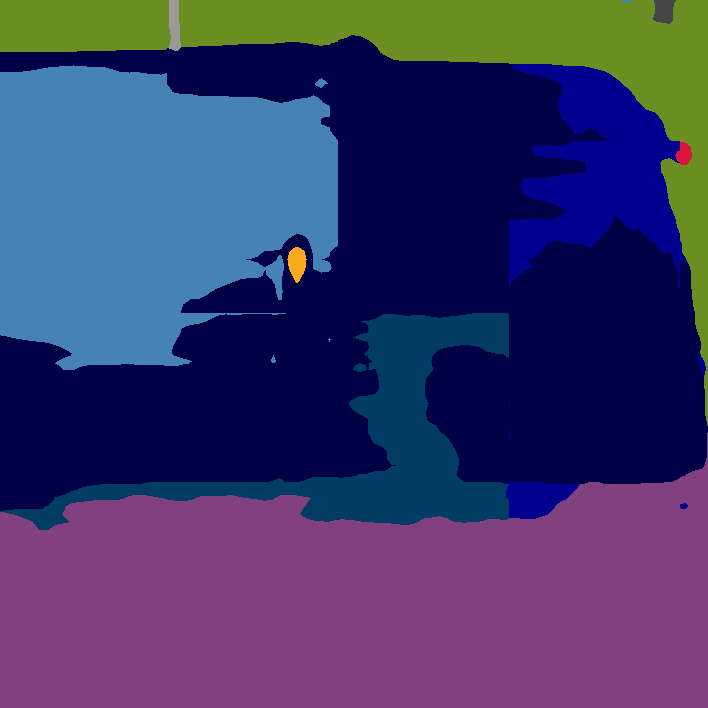}
	\end{minipage}}
	\hskip -1pt
	\subfigure{\begin{minipage}{.13\linewidth}
			\centering
			\includegraphics[width=\linewidth]{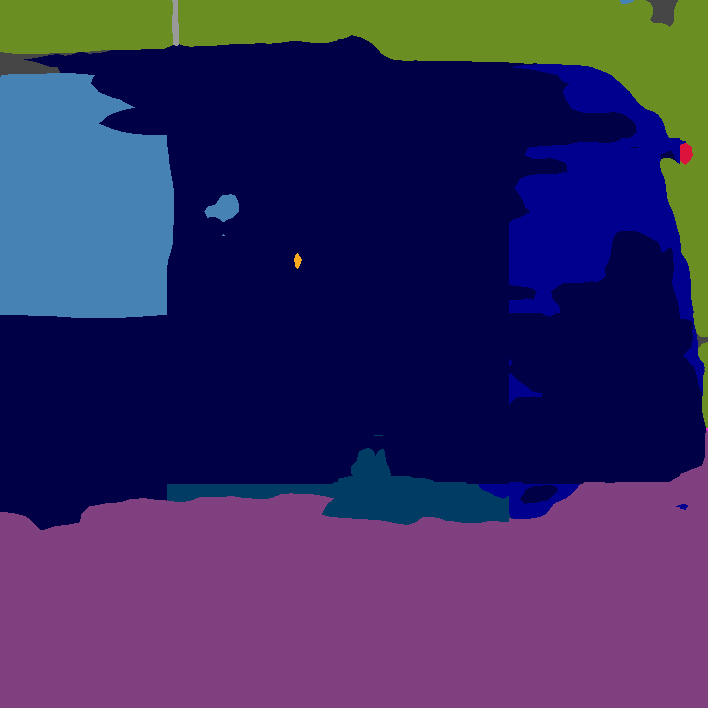}
	\end{minipage}}
	\hskip -1pt
	\subfigure{\begin{minipage}{.13\linewidth}
			\centering
			\includegraphics[width=\linewidth]{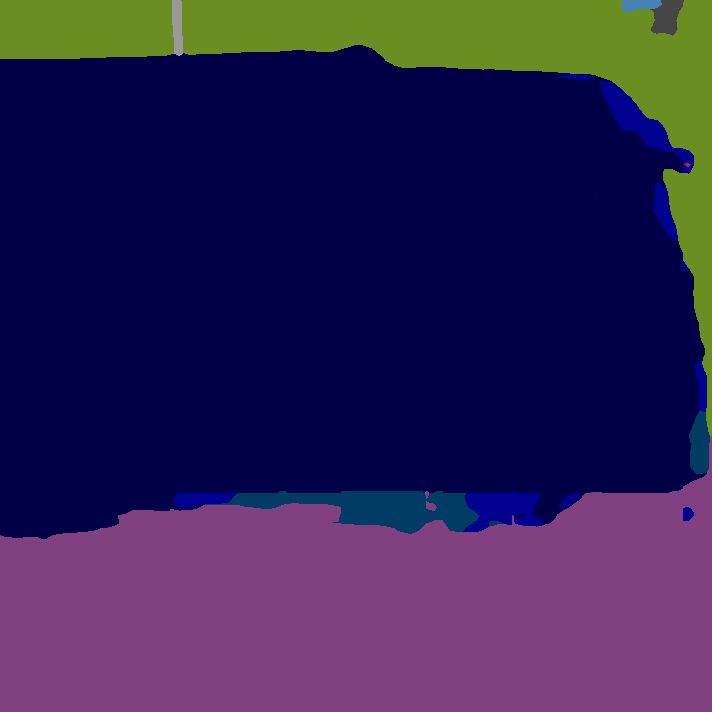}
	\end{minipage}}
	\hskip -1pt
	\subfigure{\begin{minipage}{.13\linewidth}
			\centering
			\includegraphics[width=\linewidth]{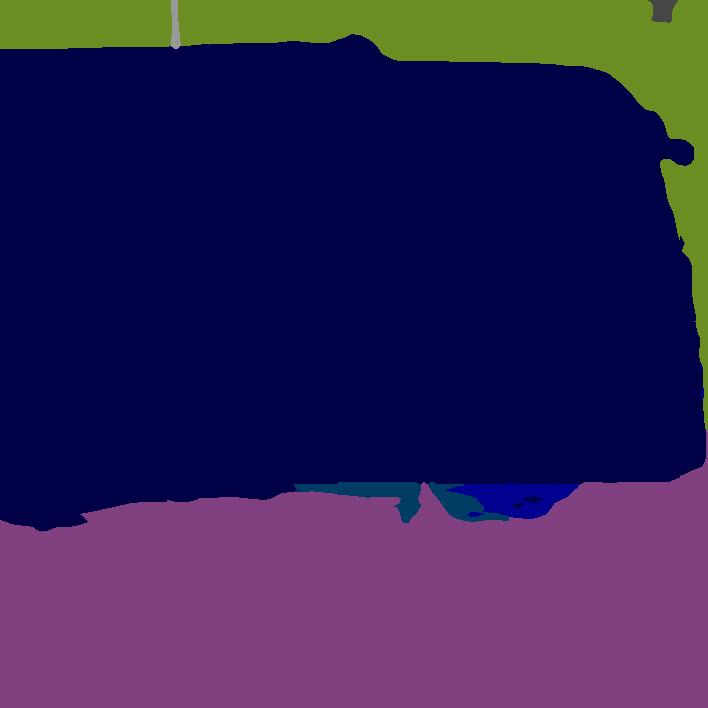}
	\end{minipage}}
	\hskip -1pt
	\subfigure{\begin{minipage}{.13\linewidth}
			\centering
			\includegraphics[width=\linewidth]{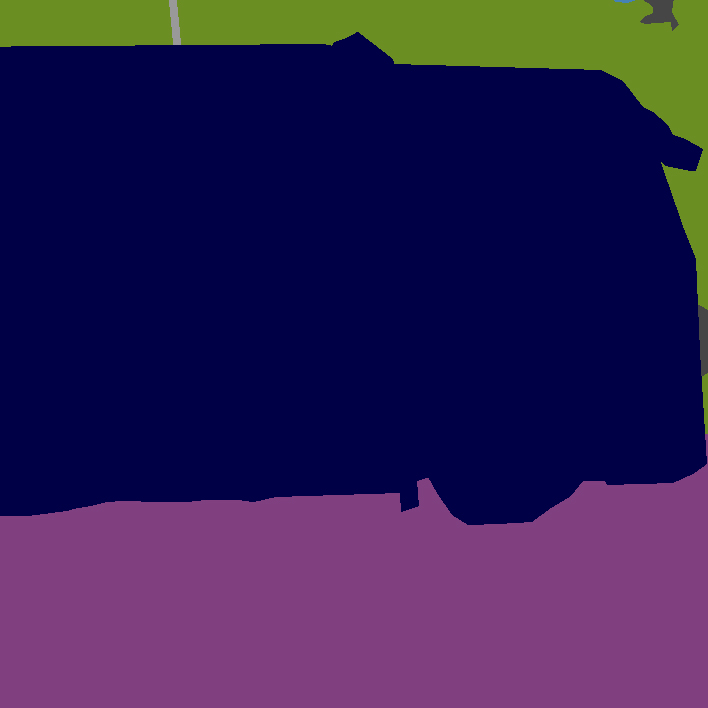}
	\end{minipage}}
	\vskip -8pt
	\subfigure{\begin{minipage}{.13\linewidth}
			\centering
			\includegraphics[width=\linewidth]{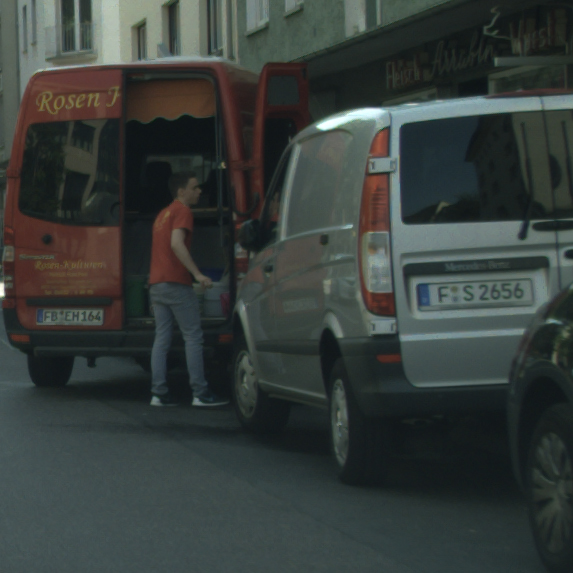}
	\end{minipage}}
	\hskip -1pt
	\subfigure{\begin{minipage}{.13\linewidth}
			\centering
			\includegraphics[width=\linewidth]{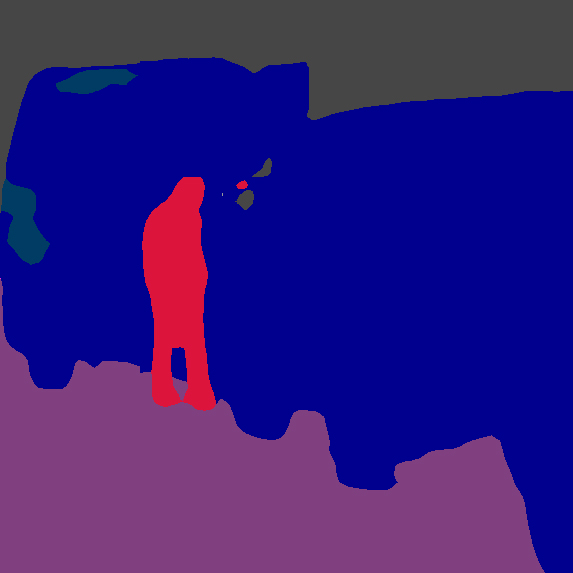}
	\end{minipage}}
	\hskip -1pt
	\subfigure{\begin{minipage}{.13\linewidth}
			\centering
			\includegraphics[width=\linewidth]{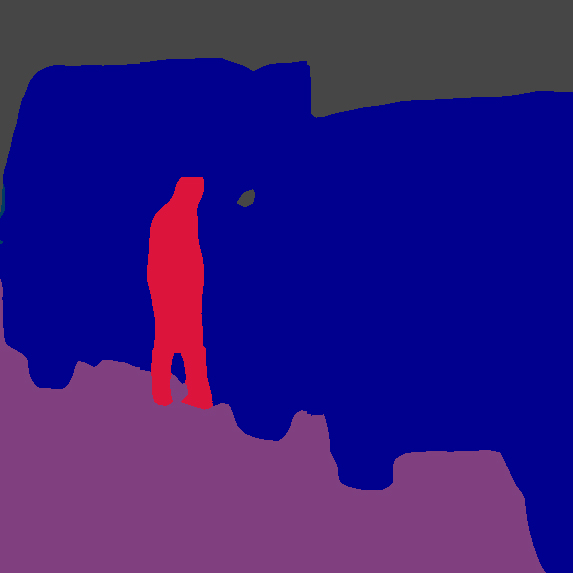}
	\end{minipage}}
	\hskip -1pt
	\subfigure{\begin{minipage}{.13\linewidth}
			\centering
			\includegraphics[width=\linewidth]{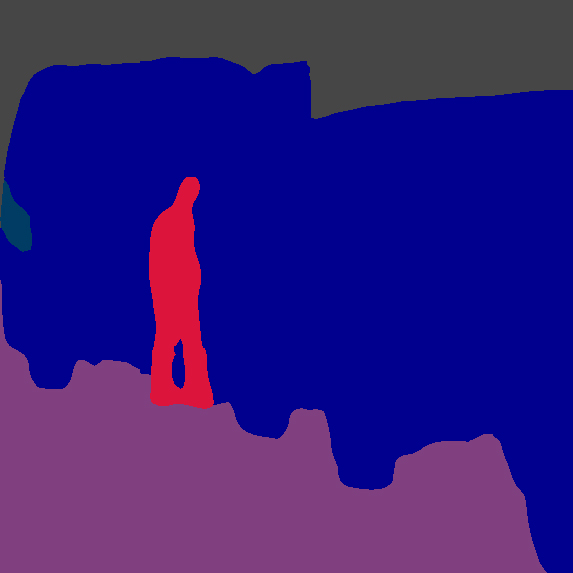}
	\end{minipage}}
	\hskip -1pt
	\subfigure{\begin{minipage}{.13\linewidth}
			\centering
			\includegraphics[width=\linewidth]{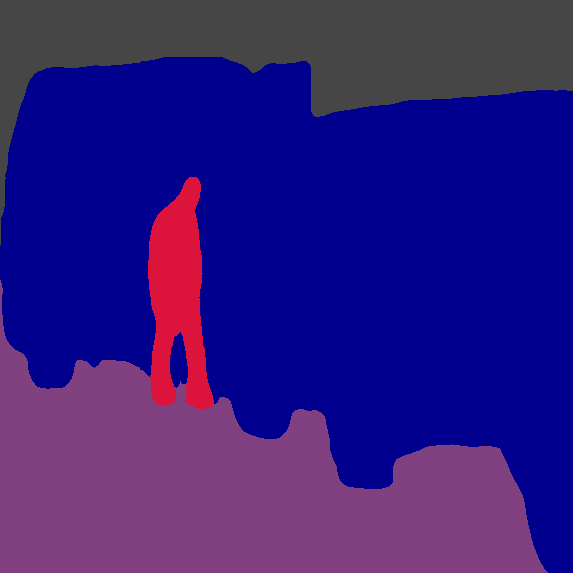}
	\end{minipage}}
	\hskip -1pt
	\subfigure{\begin{minipage}{.13\linewidth}
			\centering
			\includegraphics[width=\linewidth]{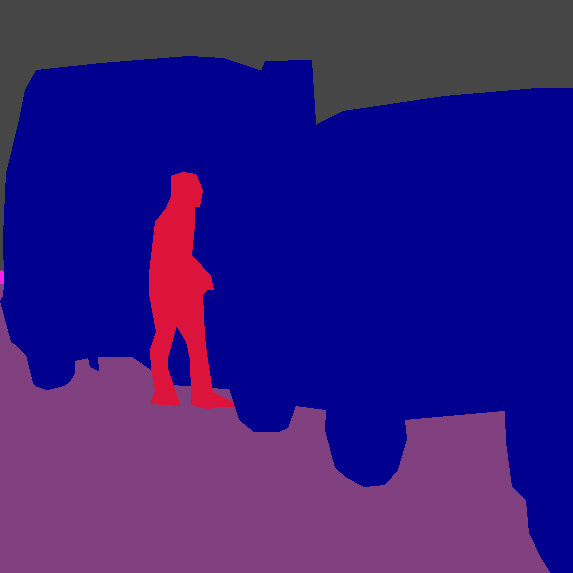}
	\end{minipage}}
	\vskip -8pt
	\setcounter{subfigure}{0}
	\subfigure[]{\begin{minipage}{.13\linewidth}
			\centering
			\includegraphics[width=\linewidth]{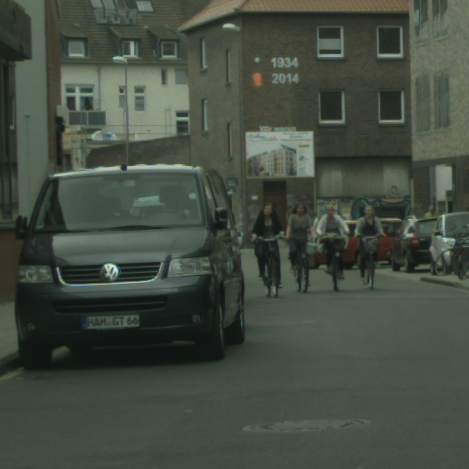}
	\end{minipage}}
	\hskip -1pt
	\subfigure[]{\begin{minipage}{.13\linewidth}
			\centering
			\includegraphics[width=\linewidth]{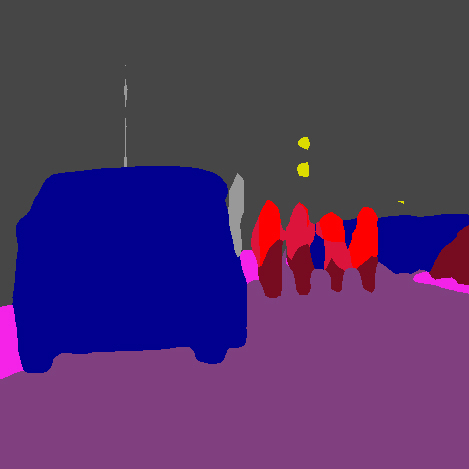}
	\end{minipage}}
	\hskip -1pt
	\subfigure[]{\begin{minipage}{.13\linewidth}
			\centering
			\includegraphics[width=\linewidth]{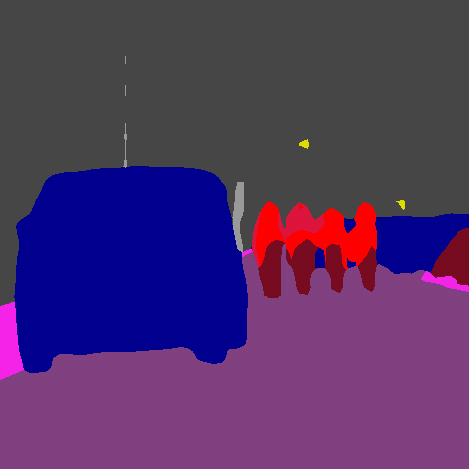}
	\end{minipage}}
	\hskip -1pt
	\subfigure[]{\begin{minipage}{.13\linewidth}
			\centering
			\includegraphics[width=\linewidth]{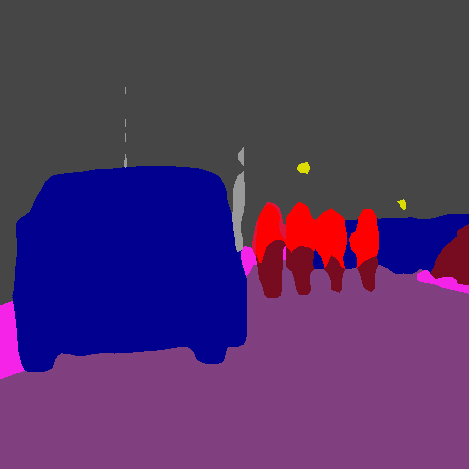}
	\end{minipage}}
	\hskip -1pt
	\subfigure[]{\begin{minipage}{.13\linewidth}
			\centering
			\includegraphics[width=\linewidth]{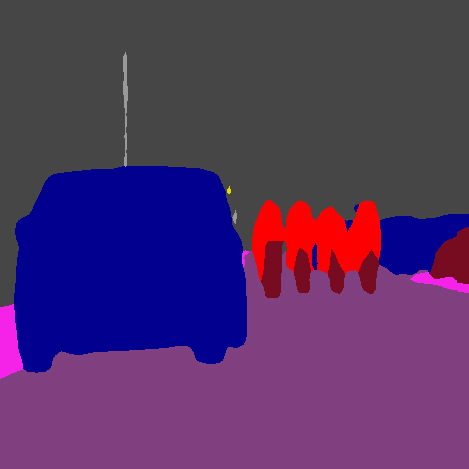}
	\end{minipage}}
	\hskip -1pt
	\subfigure[]{\begin{minipage}{.13\linewidth}
			\centering
			\includegraphics[width=\linewidth]{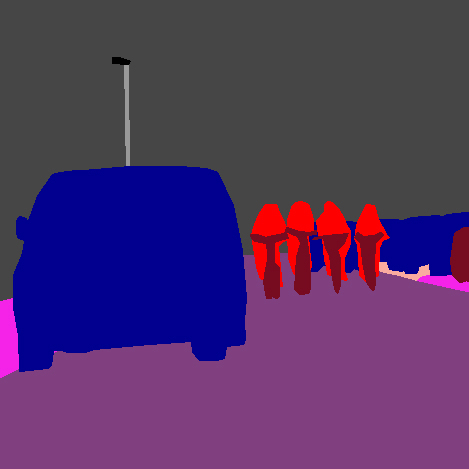}
	\end{minipage}}
	
	\caption{Visual comparison of segmentation results on the Cityscapes validation set. (a) Input image. (b) Results of attention transfer~\cite{Zagoruyko2017AT}. (c) Results of FitNet~\cite{adriana2015fitnets}.  (d) Results of affinity~\cite{He_2019_CVPR,Liu_2019_CVPR,liu2020structured}. (e) Results of the PSD module. (f) Ground truth. Best viewed in color.}
	\label{fig:com1}
\end{figure*}

\begin{table}[t]
	\begin{center}
		\caption{Effectiveness of the proposed Double Similarity Distillation on two student networks: MobileNetV2 and ResNet-18. PSD represents the pixel-wise similarity distillation module. CSD represents the category-wise similarity distillation module. Results are reported on the Cityscapes validation set.}
		\label{table:ablation1}
		\begin{adjustbox}{width=1\linewidth}
			\setlength{\tabcolsep}{6mm}{
				\begin{tabular}{@{\hspace{2.2mm}}lccc@{\hspace{2.2mm}}}
					\toprule
					Method & PSD & CSD & mIoU (\%)\\
					\midrule
					T:ResNet-101 & & & 78.23\\
					\midrule
					S1:MobileNetV2 & n/a & n/a &67.58\\
					S1:MobileNetV2 & $\surd$ & $\times$ &69.48\\
					S1:MobileNetV2 & $\times$  & $\surd$ &69.03 \\
					S1:MobileNetV2 &$\surd$ & $\surd$ & \textbf{71.25}\\
					\midrule
					S2:ResNet-18 & n/a & n/a &69.42\\
					S2:ResNet-18 & $\surd$ & $\times$ &72.24\\
					S2:ResNet-18 & $\times$ & $\surd$ &71.75 \\
					S2:ResNet-18 & $\surd$ & $\surd$ & \textbf{73.21}\\
					\bottomrule
			\end{tabular}}
		\end{adjustbox}
	\end{center}
\end{table}

\begin{table}[t]
	\begin{center}
		\caption{Ablation for our PSD module on the Cityscapes validation set. RA represents the residual attention map. $\mathcal{I}$ represents the set of selected feature maps. $A^b$, $A^h$, and $A^l$ denote the outputs of the backbone network, head network, and the logits layer.}
		\label{table:ablation2}
		\begin{adjustbox}{width=1\linewidth}
			\setlength{\tabcolsep}{8mm}{
				\begin{tabular}{@{\hspace{2.2mm}}lcc@{\hspace{2.2mm}}}
					\toprule
					RA&$\mathcal{I}$ & mIoU (\%)\\
					\midrule
					${\rm RA}^{hb}$ & $\{A^b,A^h\}$ &  71.26\\
					${\rm RA}^{lh}$& $\{A^h,A^l\}$ &71.67\\
					${\rm RA}^{lb}$& $\{A^b,A^l\}$ &71.44\\
					${\rm RA}^{lh},{\rm RA}^{hb}$& $\{A^b,A^h,A^l\}$ & \textbf{72.24}\\
					\bottomrule
			\end{tabular}}
		\end{adjustbox}
	\end{center}
\end{table}
\begin{table}[t]
	\begin{center}
		\caption{The performance in comparison on the Cityscapes validation set with FitNet~\cite{adriana2015fitnets}, AT~\cite{Zagoruyko2017AT}, and Affinity~\cite{He_2019_CVPR,Liu_2019_CVPR,liu2020structured}.}
		\label{table:ablation4}
		\begin{adjustbox}{width=1\linewidth}
			\setlength{\tabcolsep}{16.5mm}{
				\begin{tabular}{@{\hspace{2.2mm}}lc@{\hspace{2.2mm}}}
					\toprule
					Method & mIoU (\%)\\
					\midrule
					T:ResNet-101~\cite{he2016deep} & 78.23\\
					S:ResNet-18~\cite{he2016deep} & 69.42\\
					\midrule
					S + FitNet~\cite{adriana2015fitnets} & 71.31\\
					S + AT~\cite{Zagoruyko2017AT} &  71.10\\
					S + Affinity~\cite{He_2019_CVPR,Liu_2019_CVPR,liu2020structured} & 71.58\\
					S + PSD & \textbf{72.24}\\
					\bottomrule
			\end{tabular}}
		\end{adjustbox}
	\end{center}
\end{table}

\subsubsection{Effectiveness of pixel-wise similarity distillation} 

We evaluate the effect of the PSD module with different feature maps set $\mathcal{I}$. In this work, we consider outputs of the backbone network $A^b$, head network $A^h$, and the logits layer $A^l$ because of the same spatial dimension. As shown in Table~\ref{table:ablation2}, RA means the residual attention map between the two selected feature maps (see Eq. \ref{RA}). The results show that the residual attention map between any two feature maps makes similar contributions. Furthermore, we can find that the PSD calculated through the three feature maps works better than any two feature maps. This proves that the PSD module can capture the more detailed spatial dependencies as the number of layers increases. Like CCL~\cite{ding2018context} and RefineNet~\cite{Lin_2017_CVPR}, our method can be easily extended to a large set $\mathcal{I}$, such as the middle layers from the backbone. 
This is because the PSD module does not require a large number of matrix multiplications, making its FLOPs negligible. However, considering the difference in depth, architectures, and spatial dimensions of the backbone of the teacher network and the student network, it will take a lot of time and cost to find the optimal feature maps set $\mathcal{I}$. Besides, to ensure the generalization of PSD between different networks and datasets, the feature maps of middle layers from the backbone are excluded from the PSD module.

To validate the effectiveness of the proposed PSD module, we make comparisons with variants of the pixel-wise distillation: FitNet~\cite{adriana2015fitnets}, AT~\cite{Zagoruyko2017AT}, and Affinity module~\cite{He_2019_CVPR,Liu_2019_CVPR,liu2020structured}. FitNet proposed to transfer the intermediate representations of the network and add extra learning convolution layers on student network to align the matching features when the sizes are different. AT proposed to make student network mimic the attention maps of teacher network for image-level classification problem. Affinity proposed to learn the pair-wise similarity by matrix multiplication on feature map. As shown in Table~\ref{table:ablation4}, the PSD module outperforms all the other methods. The performances of the PSD module and Affinity are better than FitNet and AT. This indicates that the methods designed for image-level classification can only bring marginal improvements when applied directly to semantic image segmentation. The pixel-wise similarity distillation is more suitable for semantic image segmentation than the feature distillation and the attention map distillation. Furthermore, compared with Affinity module, the proposed method can distill the pixel-wise similarity with low computational complexity. Therefore, we can utilize multiple layers to capture more detailed spatial dependencies for better performance gains. The qualitative segmentation results in Fig.~\ref{fig:com1} visually demonstrate the effectiveness of our proposed module. 

\begin{figure}[tbp]
	\centering
	\includegraphics[width=.9\linewidth]{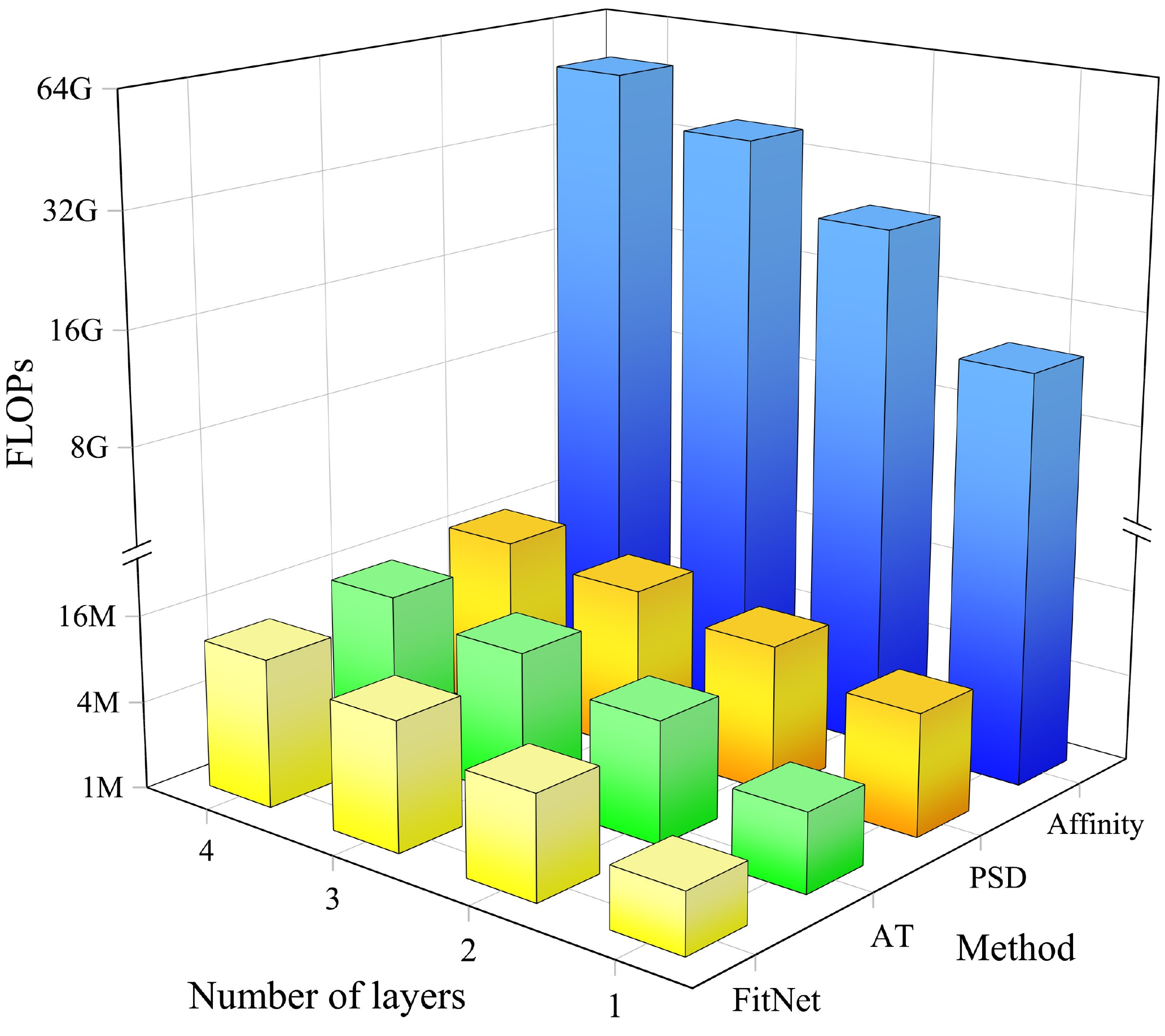}
	\caption{The FLOPs when FitNet~\cite{adriana2015fitnets}, AT~\cite{Zagoruyko2017AT}, Affinity~\cite{Liu_2019_CVPR,He_2019_CVPR,liu2020structured}, and our PSD module are applied to different numbers of layers of the network. The reshape and normalization operations are ignored and the dimensions of all the selected layers are set to $80 \times 45 \times 256$.}
	\label{fig:psd_com}
\end{figure}

\begin{figure}[t]
	\centering
	\subfigcapskip=5pt
	\subfigure[]{\begin{minipage}{.9\linewidth}
			\centering
			\includegraphics[width=\linewidth]{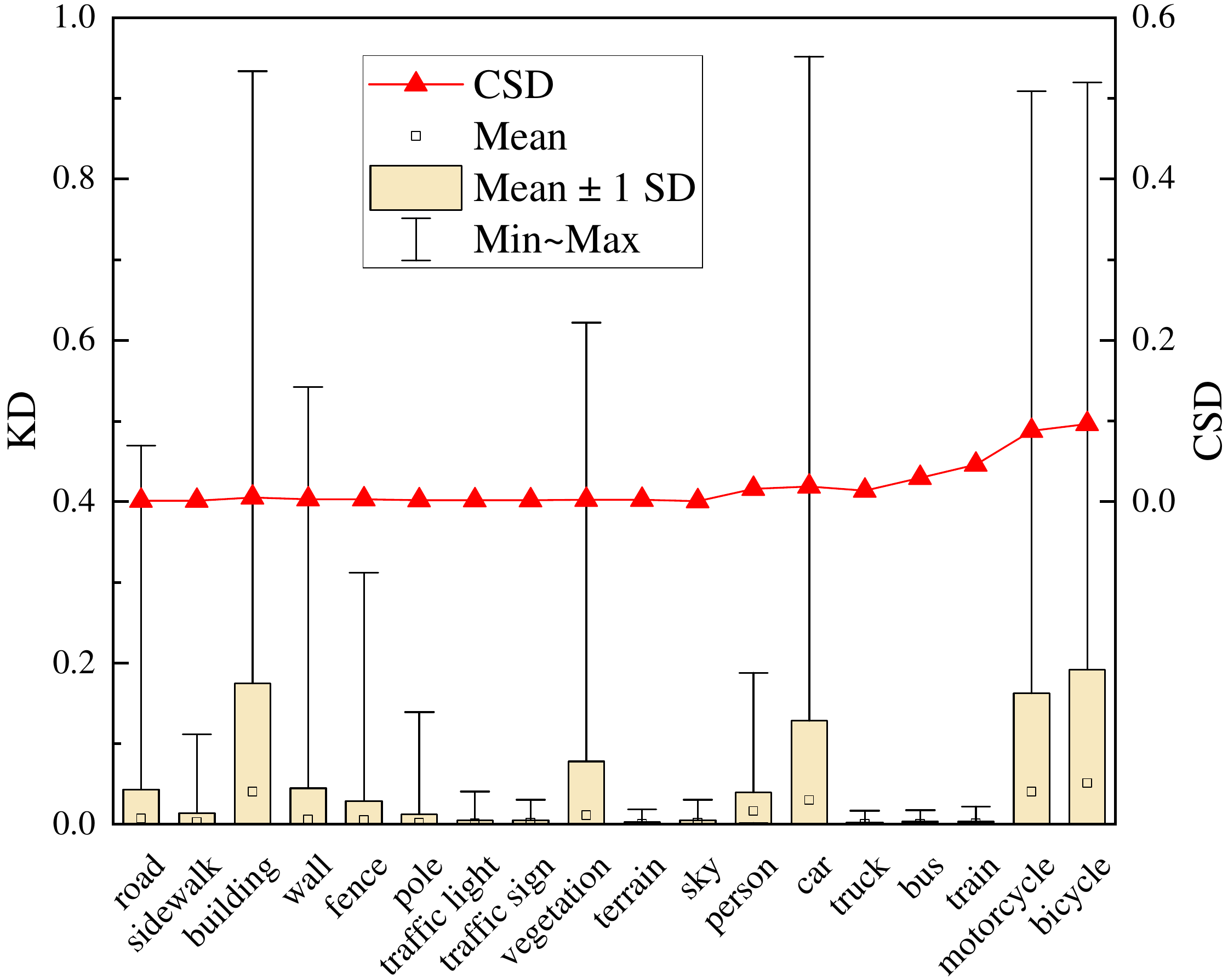}
	\end{minipage}}
	\vfill
	\subfigure[]{\begin{minipage}{.9\linewidth}
			\centering
			\includegraphics[width=\linewidth]{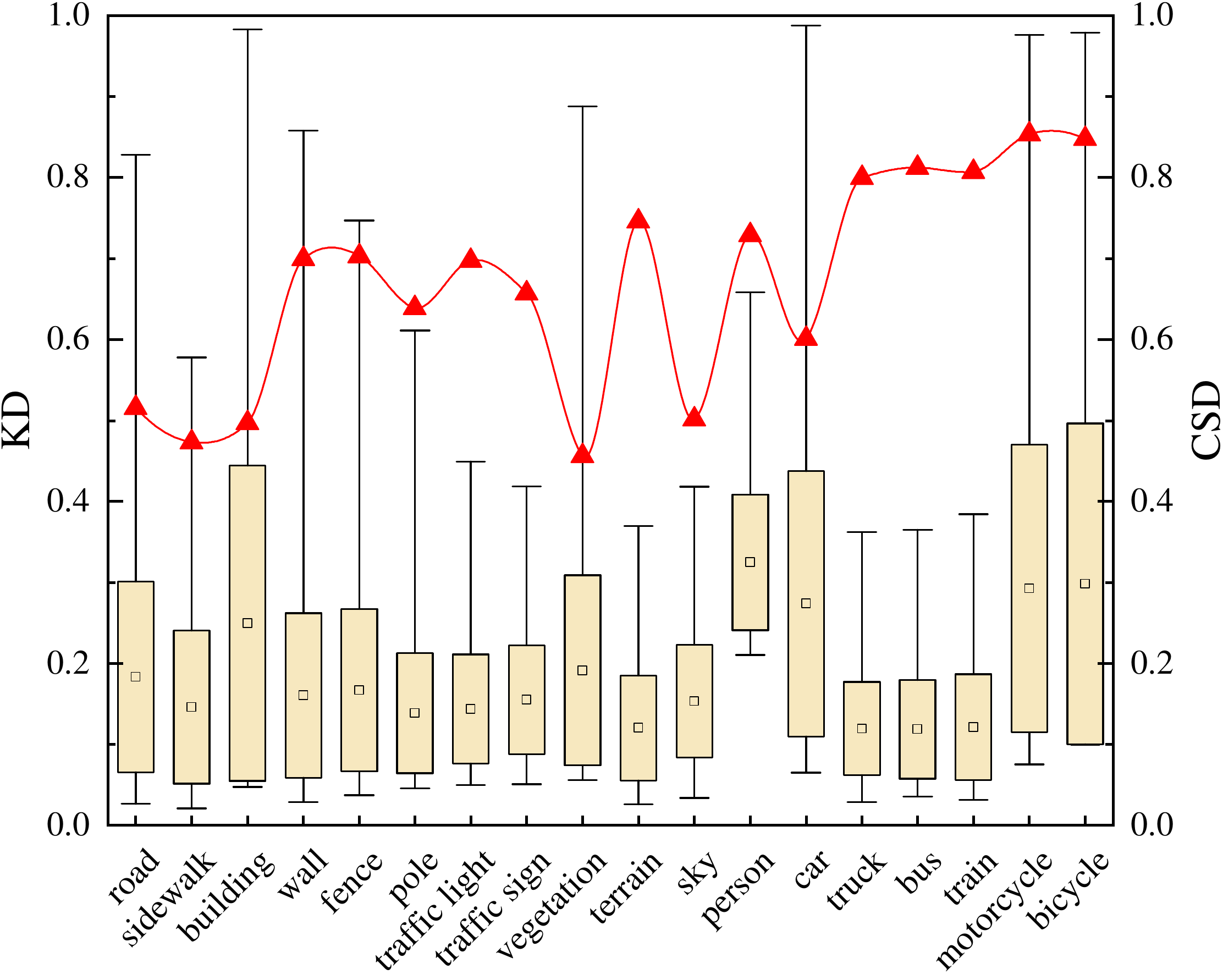}
	\end{minipage}}
	\caption{The box-plot of category correlation for the \textit{rider} class obtained by the KD method and the distribution of the global category correlation obtained by the CSD module. The upper and lower short lines represent the maximum and minimum, respectively. The small rectangle in the box and the edge of the box represent the mean and the standard deviation, respectively. The red line with the triangle symbol means the distribution obtained by the CSD module. (a) The distributions at temperatures $\tau=1$. (b) The distributions at temperatures $\tau=4$. The larger the $\tau$, the softer the distribution over classes.}
	\label{fig:na}
\end{figure}

As shown in Fig.~\ref{fig:psd_com}, to further compare the computational cost of the above four methods, we visualize the FLOPs when these methods are applied to different numbers of layers. The reshape operation and normalization operation are ignored for better understanding and comparison, and assume that the selected layers of the teacher network and the student network have the same dimensions $80 \times 45 \times 256$. It can be seen that the FLOPs of the PSD module are similar to those of FitNet~\cite{adriana2015fitnets} and AT~\cite{Zagoruyko2017AT}. Compared with the GB level FLOPs of the whole network, the value of the PSD module turns out to be negligible. However, the FLOPs of the Affinity module~\cite{Liu_2019_CVPR,liu2020structured,He_2019_CVPR} are very large. Even if it is only applied to one layer between the student network and the teacher network, the FLOPs of the Affinity module is already equivalent to or even more than the FLOPs of the compact semantic image segmentation network, which is 1800 times larger than our PSD module. Besides, the FLOPs will multiply when the Affinity is applied to multiple layers of the network. Note that we just assume that the selected layers have the same small dimensions, the true FLOPs of the Affinity module will be higher than the PSD. By comparison, the PSD module can reuse the previous layers, thereby further reducing the FLOPs. Therefore, the PSD module can be applied to multiple convolutional layers with low computational complexity to capture more detailed spatial dependencies.

\begin{table}[t]
	\begin{center}
		\caption{The performance in comparison on the Cityscapes validation set with the KD~\cite{hinton2015distilling} method at different temperature $\tau$.}
		\label{table:ablation3}
		\begin{adjustbox}{width=1\linewidth}
			\setlength{\tabcolsep}{5mm}{
				\begin{tabular}{@{\hspace{2.2mm}}lcccc@{\hspace{2.2mm}}}
					\toprule
					Method & $\tau=1$ & $\tau=2$& $\tau=4$ & $\tau=8$\\
					\midrule
					KD~\cite{hinton2015distilling} &70.56 & 70.91 & 71.21 & 71.11\\
					CSD &\textbf{71.12} & \textbf{71.33} & \textbf{71.75} & \textbf{71.19}\\
					\bottomrule
			\end{tabular}}
		\end{adjustbox}
	\end{center}
\end{table}

\begin{table}[t]
	\begin{center}
		\caption{Performance comparison on the Cityscapes test set. FLOPs is estimated for an input of $3 \times 640 \times 360$.}
		\label{table:results1}
		\begin{adjustbox}{width=1\linewidth}
			\begin{threeparttable}
				\begin{tabular}{lccc}
					\toprule
					Method & FLOPs (G) & Params (M) & mIoU (\%) \\
					\midrule
					SegNet~\cite{7803544}\tnote{\dag} & 286 & 29.5 & 57.0 \\
					ENet~\cite{paszke2016enet}\tnote{\dag}  & 3.8 & 0.4 & 58.3 \\
					ESPNet~\cite{Mehta_2018_ECCV}\tnote{\dag}  & 4.0 & 0.4 & 60.3 \\
					ICNet~\cite{Zhao_2018_ECCV_icnet}\tnote{\ddag}  & 28.3 & 26.5 & 69.5 \\
					ERFNet~\cite{romera2017erfnet}\tnote{\ddag}  & 21.0 & 2.1 & 69.7  \\
					DFANet~\cite{Li_2019_CVPR}\tnote{\ddag}  & 1.7 &7.8 & 70.3 \\
					RefineNet~\cite{Lin_2017_CVPR}\tnote{\ddag}  & 428.3 & 118.4 & 73.6  \\
					BiseNet~\cite{Yu_2018_ECCV}\tnote{\ddag}  & 55.3 & 49.0 & 74.7  \\
					SwiftNet~\cite{orsic2019defense}\tnote{\ddag}  & 218.0 & 24.7 & 76.5  \\
					PSPNet~\cite{zhao2017pyramid}\tnote{\ddag,\S}  & 255.4 & 70.4 & 78.4  \\
					\midrule
					MobileNetV2~\cite{Sandler_2018_CVPR}\tnote{\ddag} & 17.9 & 5.2 & 66.7 \\
					MobileNetV2 (ours)\tnote{\ddag} & 17.9 & 5.2 & \textbf{70.9} \\
					\midrule
					ResNet-18~\cite{he2016deep}\tnote{\ddag}  & 56.9 & 15.2 & 67.6 \\
					ResNet-18 (SKD)~\cite{Liu_2019_CVPR,liu2020structured}\tnote{\ddag}  & 56.9 & 15.2 & 71.4 \\
					ResNet-18 (ours)\tnote{\ddag}  & 56.9 & 15.2 & \textbf{72.3} \\
					\bottomrule
				\end{tabular}
				\begin{tablenotes}
					\footnotesize
					\item[\dag] Trained from scratch
					\item[\ddag] Initialized from the weights pretrained on ImageNet
					\item[\S] Tested on left-right flipping and multiple scales.
				\end{tablenotes}
			\end{threeparttable}
		\end{adjustbox}
	\end{center}
\end{table}

\begin{table*}[htbp]
	\begin{center}
		\caption{Performance comparison on the CamVid test set. The full names of the categories are Building, Tree, Sky, Car, Sign/Symbol, Road, Pedestrian, Fence, Column/Pole, Sidewalk, and Bicyclist. "-" indicates that the methods do not give the results.}
		\label{table:results5}
		\begin{adjustbox}{width=1\linewidth}
			\begin{tabular}{@{\hspace{2.2mm}}lccccccccccccc@{\hspace{2.2mm}}}
				\toprule
				Method & Params & Build. & \hspace{.1em}Tree\hspace{.1em} & \hspace{.25em}Sky\hspace{.25em} & \hspace{.25em}Car\hspace{.25em} & \hspace{.1em}Sign\hspace{.1em} & Road & Pede. & fence & pole & swalk & bicy. & mIoU (\%)\\
				\midrule
				SegNet~\cite{7803544} & 29.5 & 88.8 & 87.3 & 92.4 & 82.1 & 20.5 & 97.2 & 57.1 & 49.3 & 27.5 & 84.4 & 30.7 & 55.6\\
				BiseNet~\cite{Yu_2018_ECCV} & 27.0 & 83.0 & 75.8 & 92.0 & 83.7 & 46.5 & 94.6 & 58.8 & 53.6 & 31.9 & 81.4 & 54.0 & 68.7\\
				PSPNet (T)~\cite{zhao2017pyramid} & 70.4 & 85.5 & 77.3 & 91.2 & 90.5 & 50.6 & 95.7 & 56.7 & 51.5 & 27.2 & 85.4 & 98.0 & 73.6\\
				\midrule
				MobileNetV2~\cite{Sandler_2018_CVPR} & 5.2 & 82.7 & 75.6 & 90.4 & 85.6 & 43.2 & 93.4 & 51.2 & 41.7 & 13.1 & 78.1 & 97.4 & 68.4 \\
				MobileNetV2 (ours) & 5.2 & 82.3 & 75.8 & 90.8 & 85.7 & 46.1 & 93.6 & 55.9 & 39.4 & 23.5 & 78.6 & 97.8 & \textbf{70.0} \\
				\midrule
				ResNet-18~\cite{he2016deep} & 15.2 & 82.8 & 76.0 & 90.4 & 83.8 & 44.4 & 93.6 & 52.0 & 42.5 & 19.2 & 78.1 & 97.3 & 69.1 \\
				ResNet-18 (ours) & 15.2 & 83.8 & 76.3 & 90.6 & 86.4 & 47.9 & 94.2 & 56.9 & 38.9 & 21.0 & 80.4 & 97.7 & \textbf{70.4} \\
				\midrule
				ResNet-18 + aux~\cite{he2016deep} & 15.2 & 83.8 & 76.2 & 90.7 & 86.5 & 46.3 & 94.4 & 54.4 & 39.1 & 22.6 & 81.0 & 97.6 & 70.2 \\
				ResNet-18 + aux (SKD)~\cite{Liu_2019_CVPR,liu2020structured} & 15.2 &-&-&-&-&-&-&-&-&-&-&-&71.0\\
				ResNet-18 + aux (ours) & 15.2 & 83.9 & 77.0 & 91.1 & 87.1 & 49.0 & 95.1 & 57.0 & 39.4 & 24.9 & 83.1 & 98.0 & \textbf{71.4} \\
				\bottomrule
			\end{tabular}
		\end{adjustbox}
	\end{center}
\end{table*}

\subsubsection{Effectiveness of category-wise similarity distillation} 

In this section, we evaluate the effectiveness of our CSD module and compare with the KD~\cite{hinton2015distilling}, which proposed to improve the performance of the student network by learning the soft targets generated by the high-performance teacher networks at suitable temperature $\tau$. As can be seen in Table~\ref{table:ablation3}, to make fair comparisons, the temperature $\tau$ is set to $1$, $2$, $4$, and $8$, respectively. The larger the $\tau$, the softer the probability distribution over classes. We can find that the CSD module brings more performance gains than the KD under different temperature settings and the CSD module achieves the highest mIoU score 71.75\% when $\tau$ is set to $4$. The experiments further prove that the methods designed for other computer vision tasks can only bring sub-optimal improvements.

In order to further verify that the CSD module is more suitable for semantic segmentation than the KD~\cite{hinton2015distilling}, we compare the category correlation obtained by KD and our CSD module. For the KD method, each pixel can generate one soft category distribution. As shown in Fig.~\ref{fig:na}, we count the category correlation of the pixels belong to the \textit{rider} class in one image on Cityscapes validation set and plot the box-plot to measure the dispersion of the distributions. The upper and lower short lines represent the maximum and minimum, respectively. The small rectangle in the box and the edge of the box represent the mean and the standard deviation, respectively. The longer the box, the larger the standard deviation and the more scattered the data. It can be seen that the category correlation of pixels belonging to the same category are not uniform. It is difficult for the student network to understand the complicated and contradictory category distributions. On the contrary, the CSD module can integrate the soft category distributions for each pixel belonging to the same category and generate only one global distribution in the whole image. The global category correlation reduces the learning difficulty of the student network.

\subsection{Comparison with the State-of-the-Arts} 

In this section, we show the comparison with the state-of-the-art methods on the four publicly datasets. Tables~\ref{table:results1}-\ref{table:results4} detail the segmentation accuracy on different datasets. Fig.~\ref{fig:results} shows the segmentation results of the proposed method.

\subsubsection{Cityscapes}

We report the results and the comparisons with state-of-the-art methods on the Cityscapes test set. As shown in Table~\ref{table:results1}, the proposed method helps the compact semantic image segmentation networks achieve significant improvements. Specifically, the mIoU score is improved from 66.7\% to 70.9\% with MobileNetV2 and 67.6\% to 72.3\% with ResNet-18. Compared with SKD~\cite{Liu_2019_CVPR,liu2020structured} that combines pixel-wise, pair-wise, and holistic distillation, the proposed method based on the same network achieves better performance improvements. What's more, different with the holistic distillation that introduces the conditional generative adversarial network to align features, our method does not require any training parameters and can be trained end-to-end. Fig.~\ref{fig:results} shows several samples on Cityscapes validation set, the better segmentation results are achieved by the proposed distilled student network, which also further demonstrate the effectiveness of our method.

\begin{table}[t]
	\begin{center}
		\caption{Performance comparison on Pascal VOC 2012 validation set.}
		\label{table:results3}
		\begin{adjustbox}{width=1\linewidth}
			\setlength{\tabcolsep}{5.5mm}{
				\begin{threeparttable}
					\begin{tabular}{@{\hspace{2.2mm}}lcc@{\hspace{2.2mm}}}
						\toprule
						Method & Params (M) & mIoU (\%) \\
						\midrule
						Dilated-8~\cite{yu2015multi}\tnote{\dag} & 141.1 &73.9\\
						DeepLab-v2~\cite{chen2017deeplab}\tnote{\dag} & 44.5 &76.3\\
						PSPNet (T)~\cite{zhao2017pyramid} & 70.4 & 79.1\\
						\midrule
						MD~\cite{xie2018improving}\tnote{\dag} & 14.4 & 67.3\\
						MDE~\cite{xie2018improving}\tnote{\dag,\ddag}&14.4 & 69.6\\
						PFS~\cite{shan2019distilling}\tnote{\dag} &14.4 & 72.9\\
						\midrule
						MobileNetV2~\cite{Sandler_2018_CVPR} & 5.2 & 70.6 \\
						MobileNetV2 (KA)~\cite{He_2019_CVPR}\tnote{\dag} & 5.2 & 72.5 \\
						MobileNetV2 (ours) & 5.2 & \textbf{72.9} \\
						\midrule
						ResNet-18~\cite{he2016deep} & 15.2 & 71.9 \\
						ResNet-18 (ours) & 15.2 & \textbf{73.7} \\
						\bottomrule
					\end{tabular}
					\begin{tablenotes}
						\footnotesize
						\item[\dag] pre-trained on COCO dataset.
						\item[\ddag] used extra 10k unlabeled images.
					\end{tablenotes}
			\end{threeparttable}}
		\end{adjustbox}
	\end{center}
\end{table}

\subsubsection{CamVid}

The accuracy results on the test set are shown in Table~\ref{table:results5}. Although the compact student networks can obtain nearly segmentation accuracy to the cumbersome teacher networks due to the simplicity of the dataset, the DSD framework can still improve the performance of the student networks. Our method boosts the accuracy of ResNet-18 and MobileNetV2 by 1.3 and 1.6 points, respectively. Besides, considering that the auxiliary loss can help the semantic segmentation model learn more generalized representations, we add auxiliary loss to ResNet-18 and ours for a fair comparison. We can find that the performance of the ResNet-18 has increased from 69.1\% to 70.2\% due to the help of the auxiliary loss. However, our method can still further improve the performance of the ResNet-18 to 71.4\%. Compared with SKD, our method achieves better segmentation accuracy.

\begin{table}[t]
	\begin{center}
		\centering
		\caption{Performance comparison on the ADE20K validation set and test set. "-" indicates that the methods do not give the corresponding results.}
		\label{table:results4}
		\begin{adjustbox}{width=1\linewidth}
			\setlength{\tabcolsep}{1.8mm}{
				\begin{tabular}{@{\hspace{2.2mm}}lccc@{\hspace{2.2mm}}}
					\toprule
					\multirow{2}{*}{Method} & \multirow{2}{*}{Params (M)}&  \multicolumn{2}{c}{mIoU (\%) / Pixel Acc. (\%)}\\
					\cline{3-4}
					& &\multicolumn{2}{c}{val / test}\\
					\midrule
					SegNet~\cite{7803544} & 29.5 & 21.6 / 71.0 & 17.54 / 64.0\\
					FCN-8s~\cite{long2015fully} & 134.5 &29.4 / 71.3 & 24.8 / 64.8\\
					DilatedNet-50~\cite{Xiao_2018_ECCV} & 62.7 & 32.3 / 73.6 & 25.9 / 65.4\\
					PSPNet (T)~\cite{zhao2017pyramid} & 70.4 &43.1 / 81.1 &  44.2 / 81.7\\
					\midrule
					MobileNetV2~\cite{Sandler_2018_CVPR} & 5.2 & 34.4 / 76.9 & 28.0 / 68.8\\
					MobileNetV2 (ours) & 5.2 & \textbf{37.6} / \textbf{77.8} & \textbf{30.6} / \textbf{70.0}\\
					\midrule
					ResNet-18~\cite{he2016deep} & 15.2 & 33.8 / 76.1 &  27.1 / 68.4\\
					ResNet-18 (SKD)~\cite{Liu_2019_CVPR,liu2020structured} & 15.2 & 36.6 / 77.8 & - / -\\
					ResNet-18 (ours) & 15.2 & \textbf{38.0} / \textbf{78.1} & \textbf{31.3} / \textbf{70.0}\\
					\bottomrule
			\end{tabular}}
		\end{adjustbox}
	\end{center}
\end{table}

\subsubsection{Pascal VOC 2012}

As shown in Table~\ref{table:results3}, our method gets mIoU score 73.7\% for ResNet-18, which is a similar segmentation accuracy compared with Dilated-8~\cite{yu2015multi}. However, our method only needs much small network parameter. For MobileNetV2, the mIoU score is improved from 70.6\% to 72.9\%. Compared with the KA~\cite{He_2019_CVPR}, our method does not need extra training images from the COCO dataset. More importantly, KA~\cite{He_2019_CVPR} needs to optimize the auto-encoder network in advance. Besides, compared with MDE~\cite{xie2018improving} and PFS~\cite{shan2019distilling} that based on the MobileNet~\cite{howard2017mobilenets} and ASPP~\cite{chen2017deeplab} module, our method can achieve the equivalent precision with small model size. Note that these methods also need pre-trained on COCO dataset. Furthermore, the proposed method dose not need to add a specially designed module (PFS~\cite{shan2019distilling}) to retrain the teacher network, and can be applied between any existing student and teacher networks. The qualitative segmentation results on the Pascal VOC 2012 validation set can be seen in Fig.~\ref{fig:results}.

\subsubsection{ADE20K}

\begin{figure}[t]
	\centering
	\subfigcapskip=5pt
	\subfigure{\begin{minipage}{.245\linewidth}
			\centering
			\includegraphics[width=\linewidth,height=.8\linewidth]{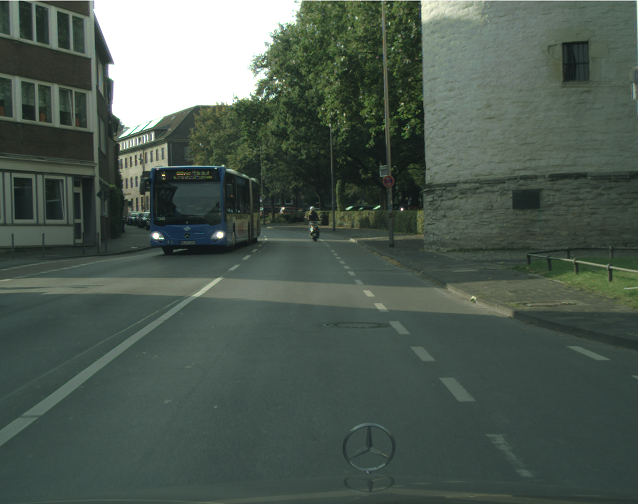}
	\end{minipage}}
	\hskip -2pt
	\subfigure{\begin{minipage}{.245\linewidth}
			\centering
			\includegraphics[width=\linewidth,height=.8\linewidth]{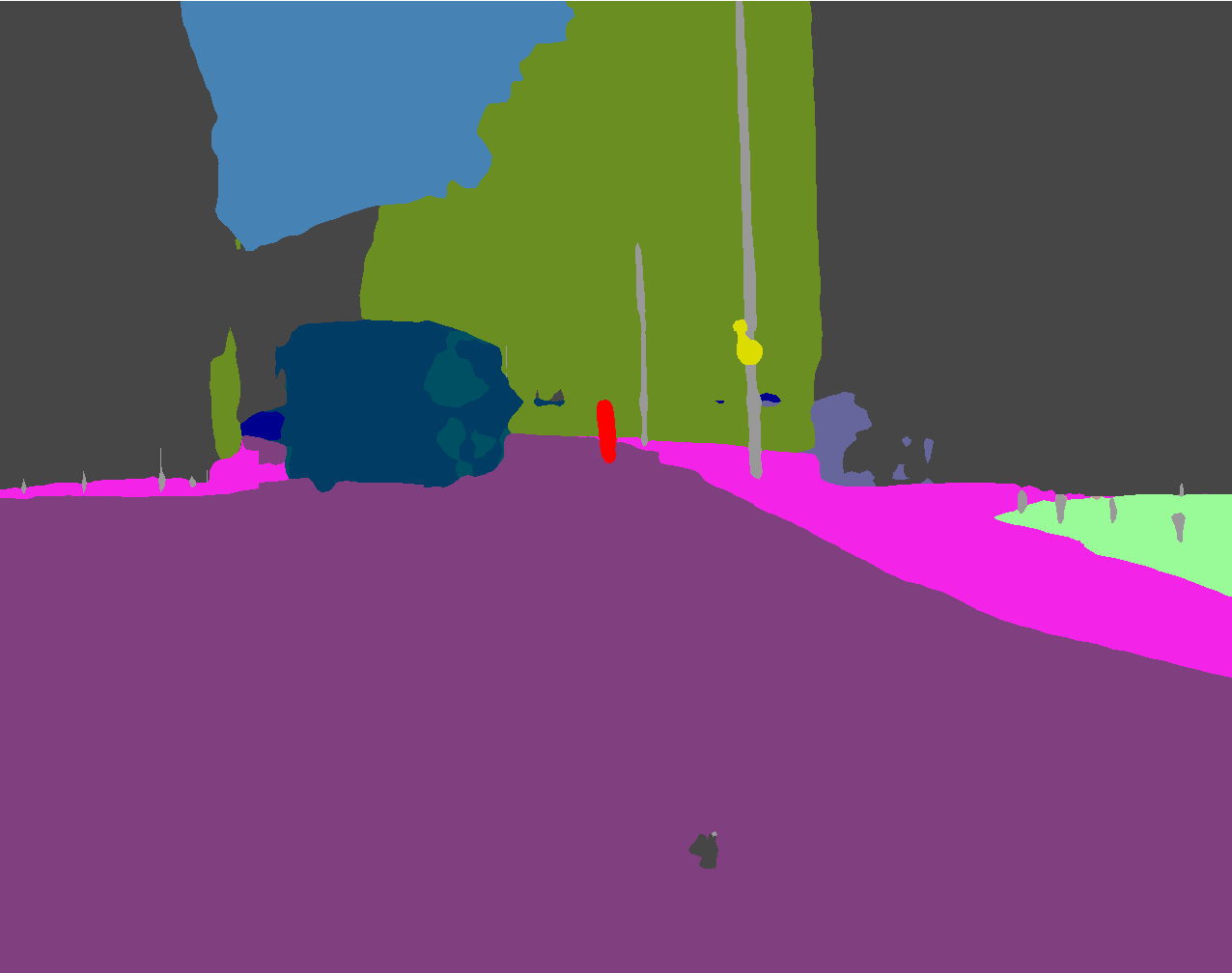}
	\end{minipage}}
	\hskip -2pt
	\subfigure{\begin{minipage}{.245\linewidth}
			\centering
			\includegraphics[width=\linewidth,height=.8\linewidth]{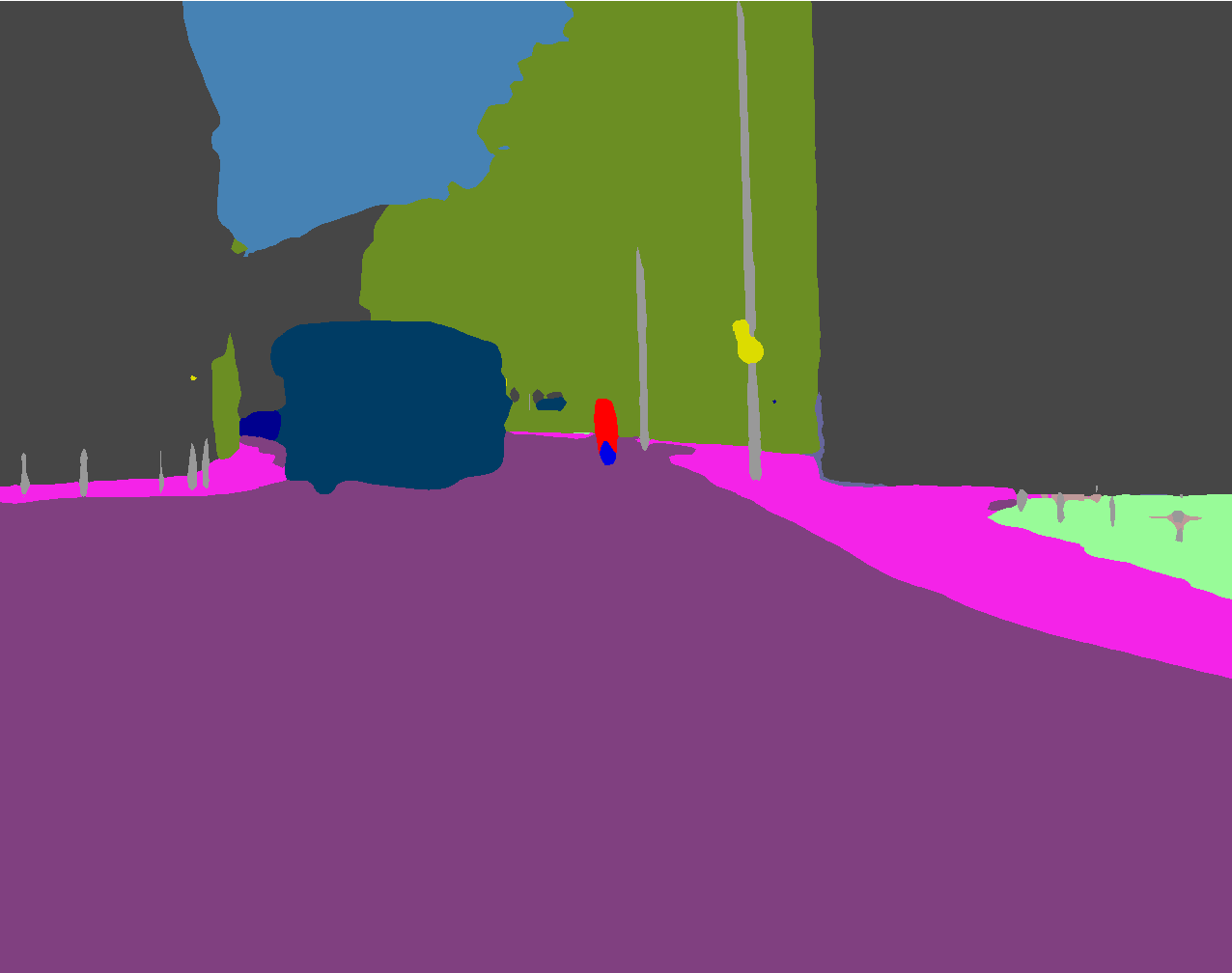}
	\end{minipage}}
	\hskip -2pt
	\subfigure{\begin{minipage}{.245\linewidth}
			\centering
			\includegraphics[width=\linewidth,height=.8\linewidth]{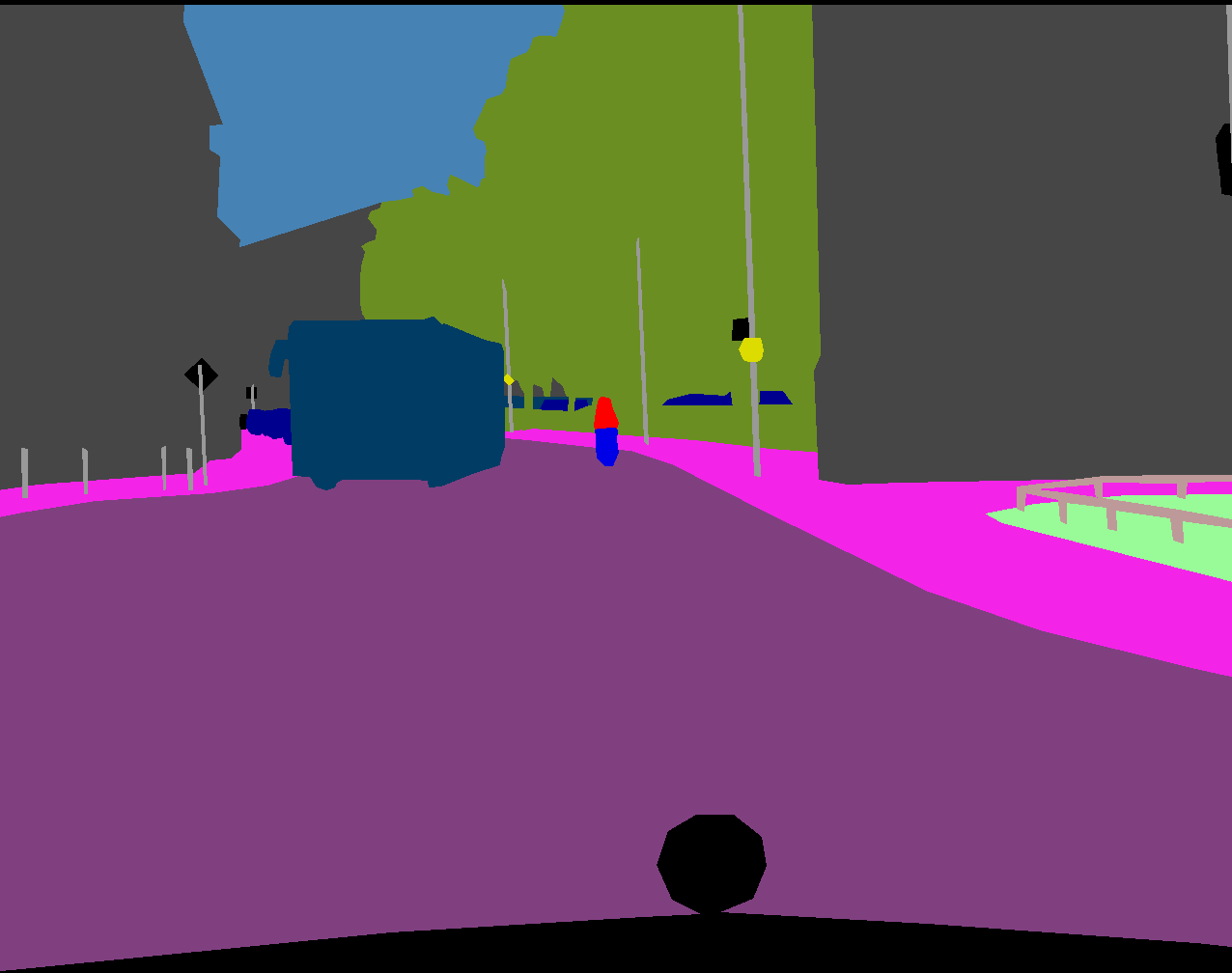}
	\end{minipage}}
	\vskip -8pt
	\subfigure{\begin{minipage}{.245\linewidth}
			\centering
			\includegraphics[width=\linewidth,height=.8\linewidth]{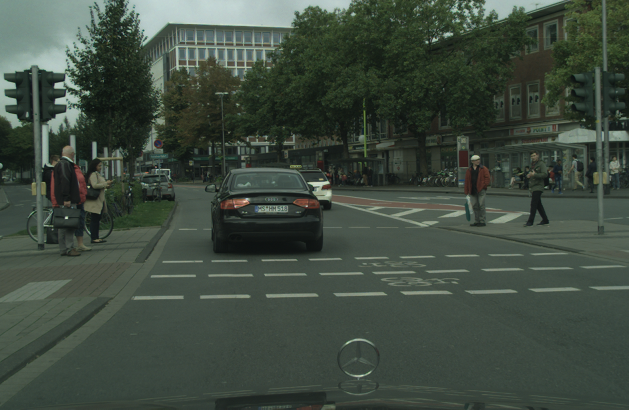}
	\end{minipage}}
	\hskip -2pt
	\subfigure{\begin{minipage}{.245\linewidth}
			\centering
			\includegraphics[width=\linewidth,height=.8\linewidth]{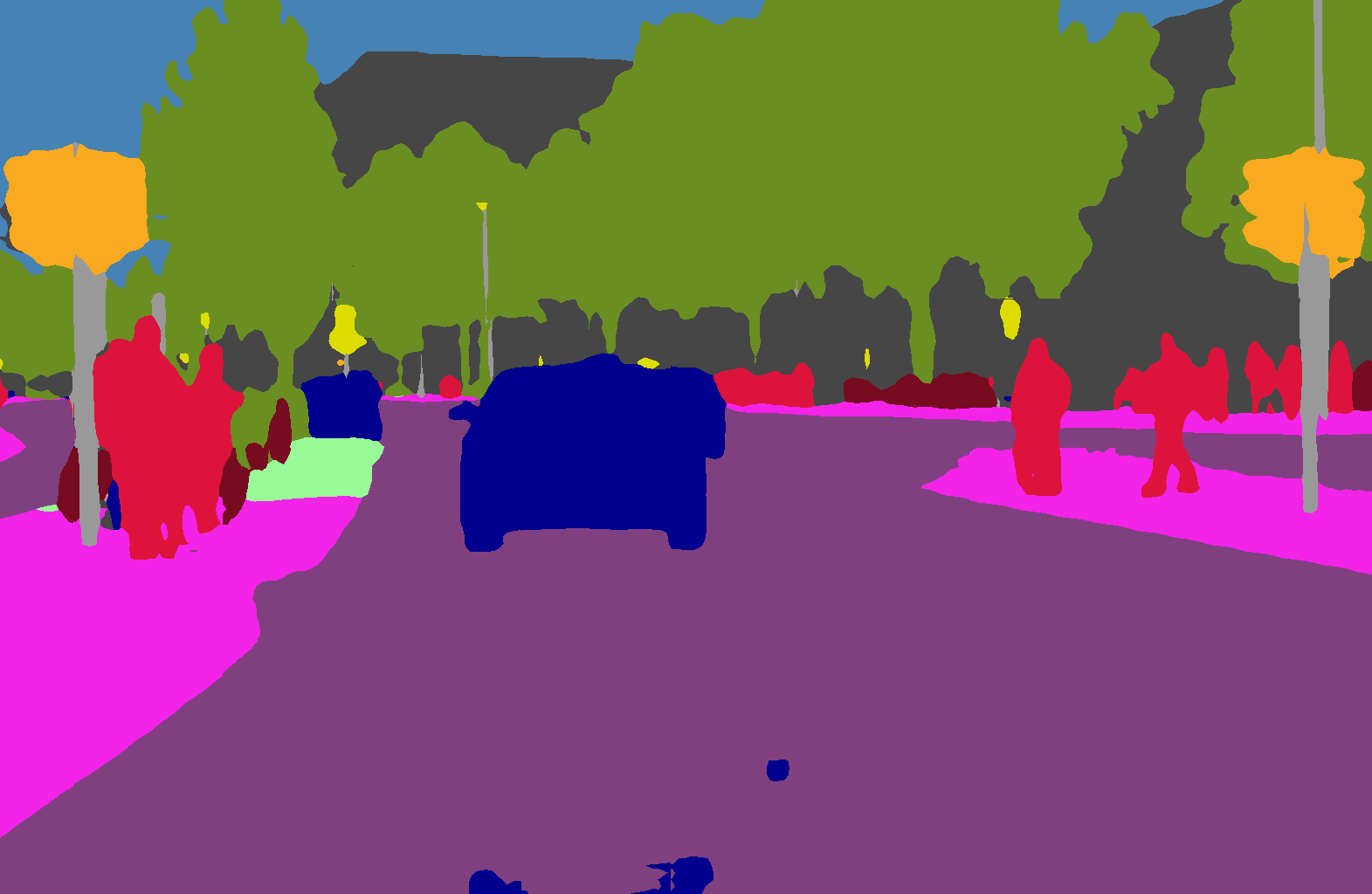}
	\end{minipage}}
	\hskip -2pt
	\subfigure{\begin{minipage}{.245\linewidth}
			\centering
			\includegraphics[width=\linewidth,height=.8\linewidth]{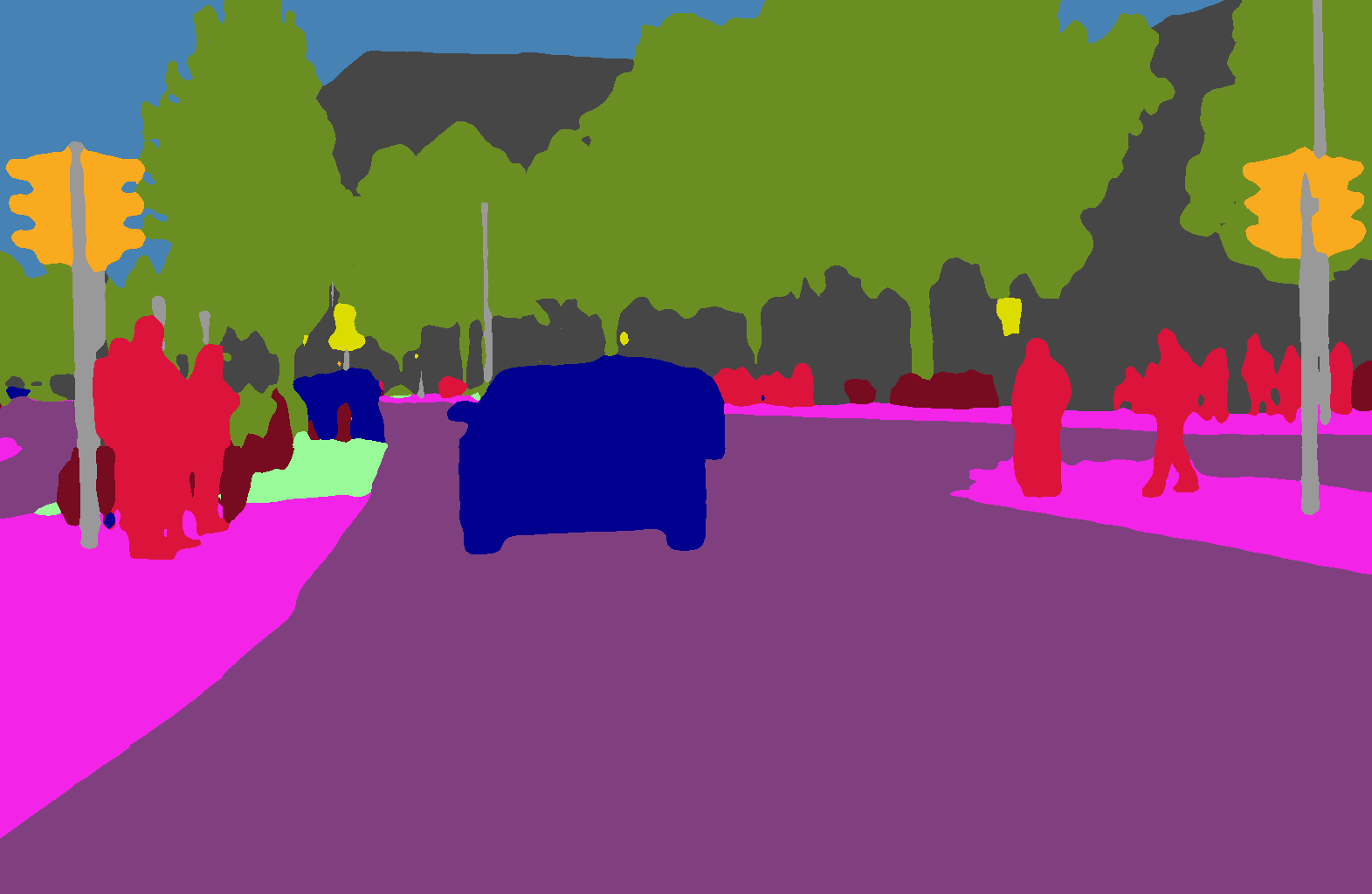}
	\end{minipage}}
	\hskip -2pt
	\subfigure{\begin{minipage}{.245\linewidth}
			\centering
			\includegraphics[width=\linewidth,height=.8\linewidth]{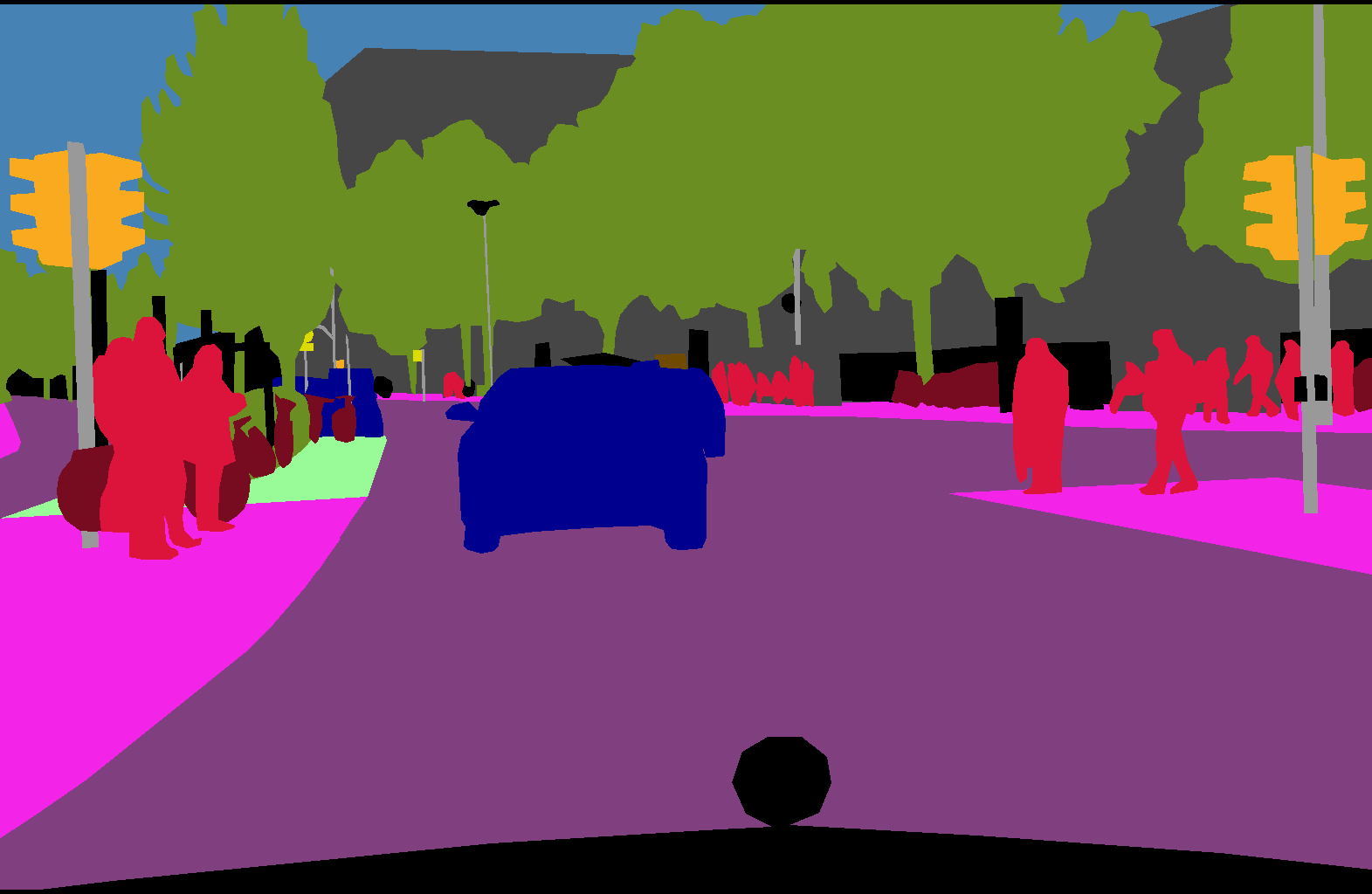}
	\end{minipage}}
	\vskip -8pt
	\subfigure{\begin{minipage}{.245\linewidth}
			\centering
			\includegraphics[width=\linewidth,height=.8\linewidth]{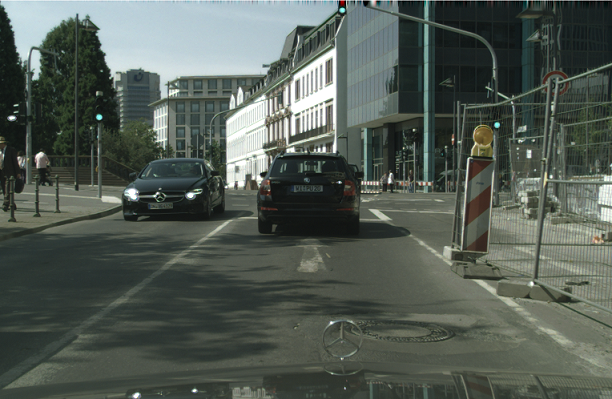}
	\end{minipage}}
	\hskip -2pt
	\subfigure{\begin{minipage}{.245\linewidth}
			\centering
			\includegraphics[width=\linewidth,height=.8\linewidth]{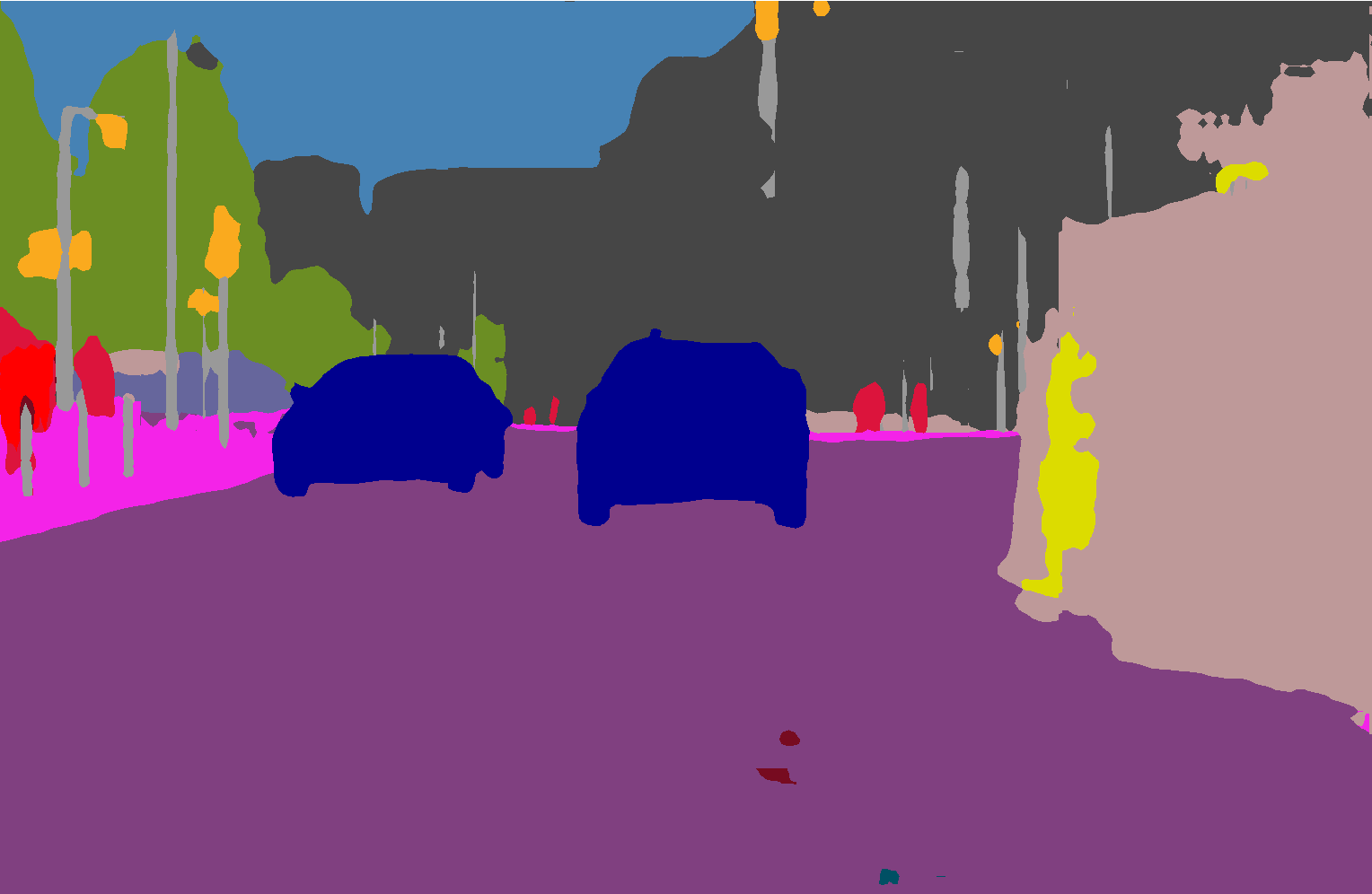}
	\end{minipage}}
	\hskip -2pt
	\subfigure{\begin{minipage}{.245\linewidth}
			\centering
			\includegraphics[width=\linewidth,height=.8\linewidth]{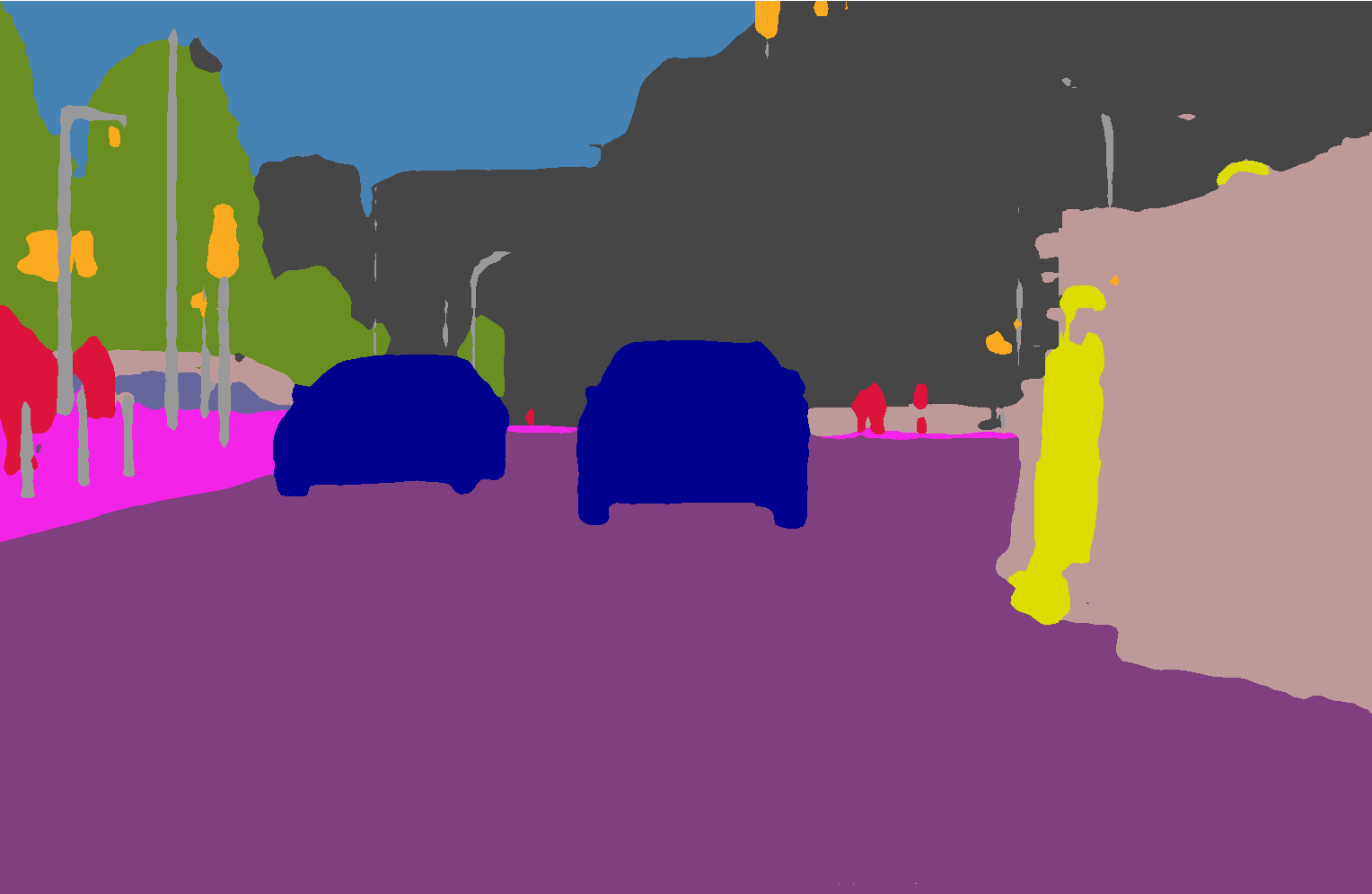}
	\end{minipage}}
	\hskip -2pt
	\subfigure{\begin{minipage}{.245\linewidth}
			\centering
			\includegraphics[width=\linewidth,height=.8\linewidth]{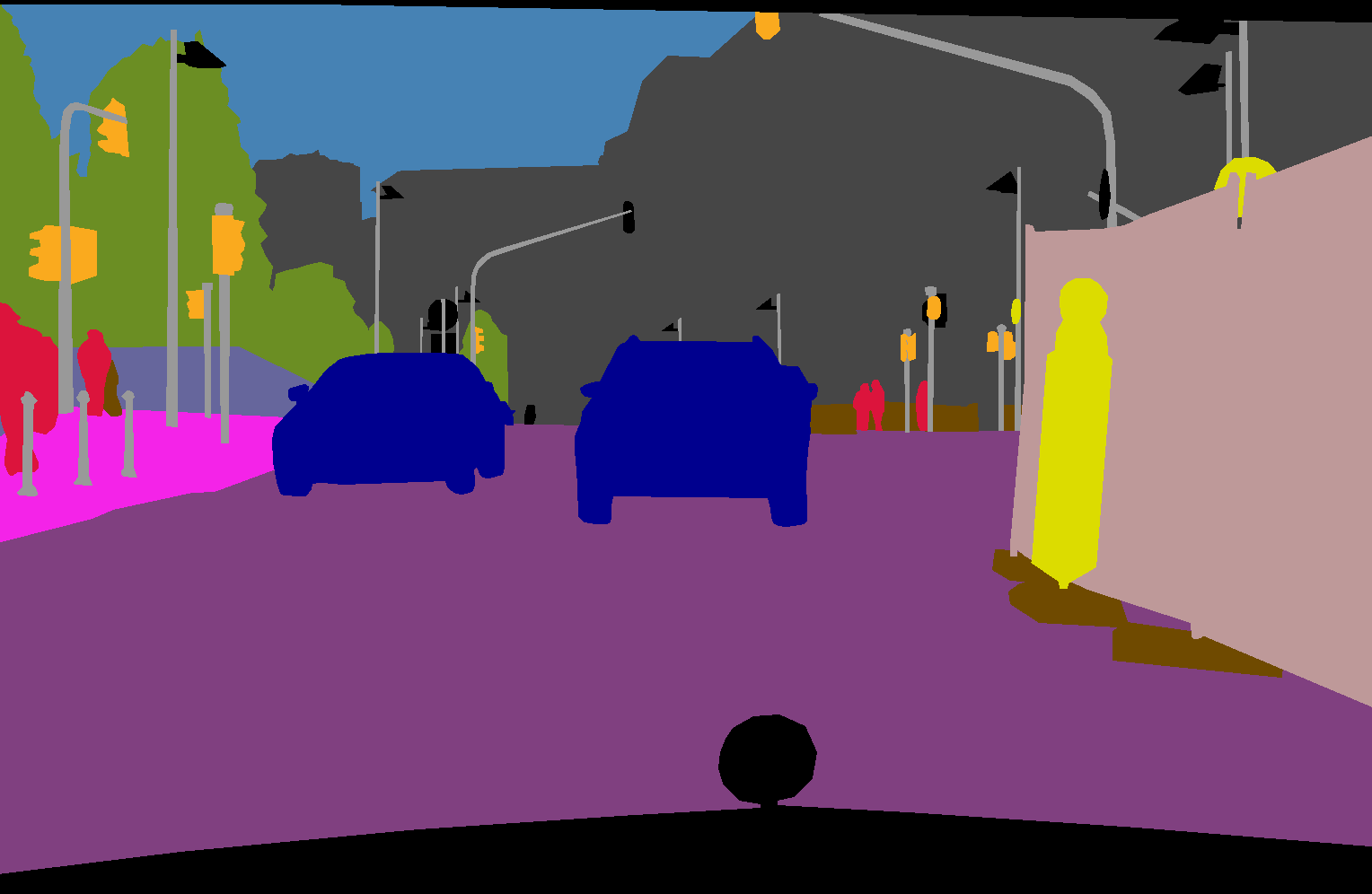}
	\end{minipage}}
	\vskip -2.5pt
	{\color{red}\rule{\linewidth}{2pt}}
	\vskip -2.5pt
	\subfigure{\begin{minipage}{.245\linewidth}
			\centering
			\includegraphics[width=\linewidth,height=.8\linewidth]{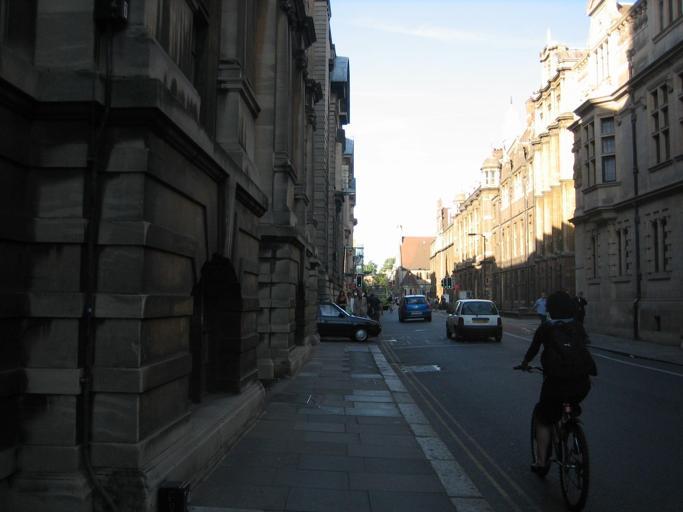}
	\end{minipage}}
	\hskip -2pt
	\subfigure{\begin{minipage}{.245\linewidth}
			\centering
			\includegraphics[width=\linewidth,height=.8\linewidth]{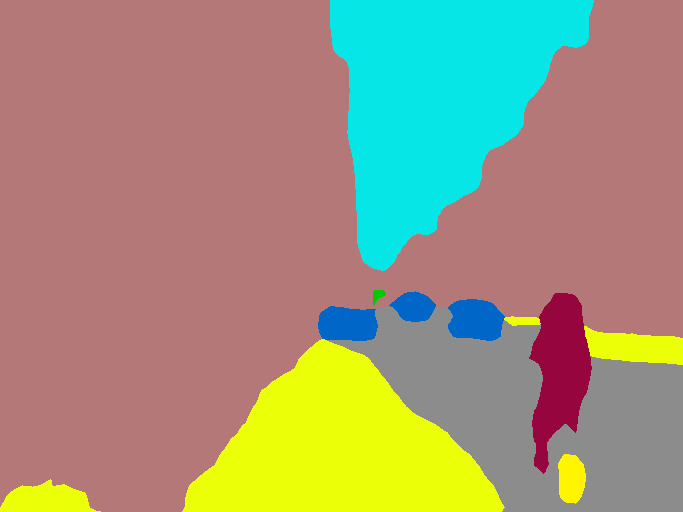}
	\end{minipage}}
	\hskip -2pt
	\subfigure{\begin{minipage}{.245\linewidth}
			\centering
			\includegraphics[width=\linewidth,height=.8\linewidth]{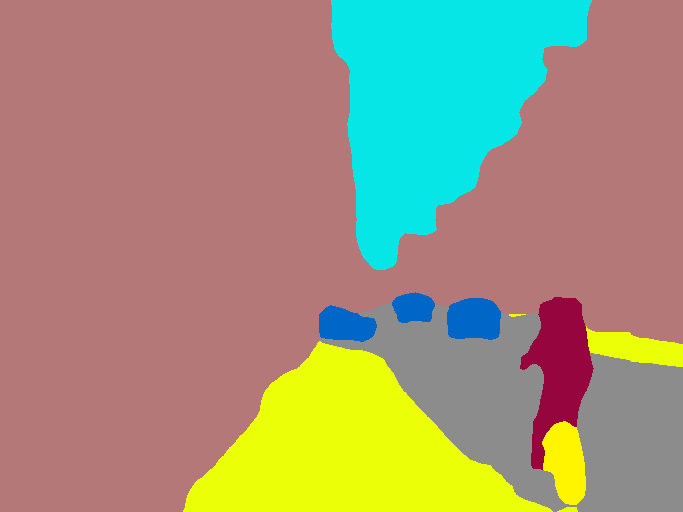}
	\end{minipage}}
	\hskip -2pt
	\subfigure{\begin{minipage}{.245\linewidth}
			\centering
			\includegraphics[width=\linewidth,height=.8\linewidth]{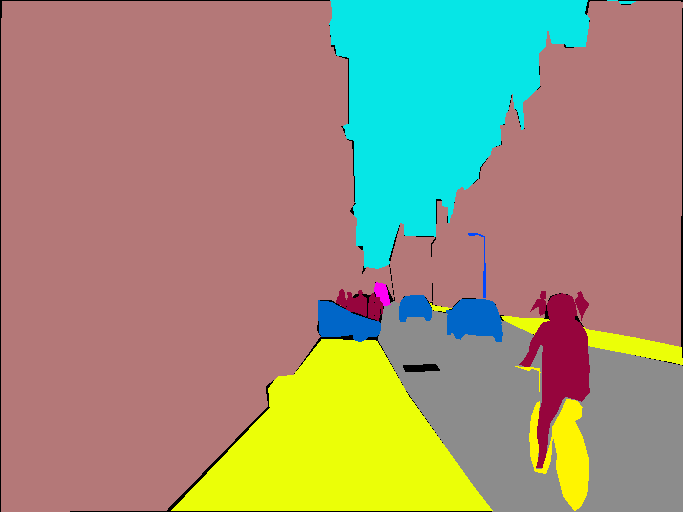}
	\end{minipage}}
	\vskip -8pt
	\subfigure{\begin{minipage}{.245\linewidth}
			\centering
			\includegraphics[width=\linewidth,height=.8\linewidth]{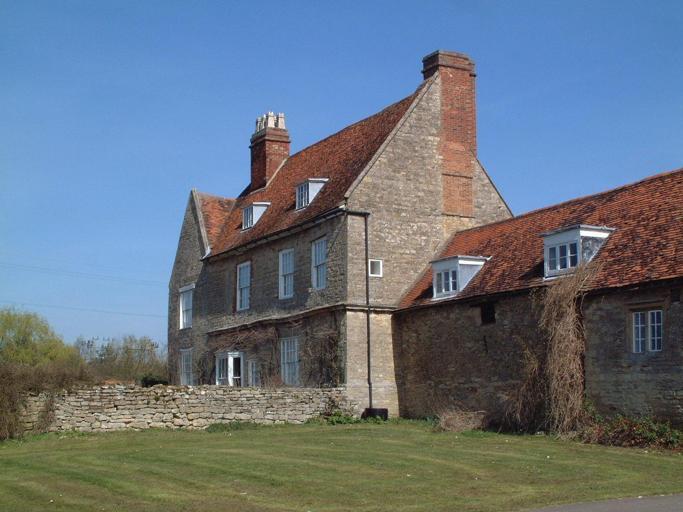}
	\end{minipage}}
	\hskip -2pt
	\subfigure{\begin{minipage}{.245\linewidth}
			\centering
			\includegraphics[width=\linewidth,height=.8\linewidth]{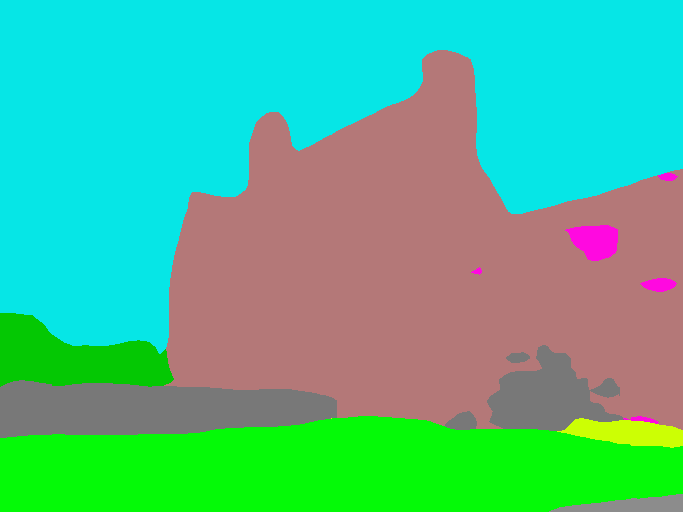}
	\end{minipage}}
	\hskip -2pt
	\subfigure{\begin{minipage}{.245\linewidth}
			\centering
			\includegraphics[width=\linewidth,height=.8\linewidth]{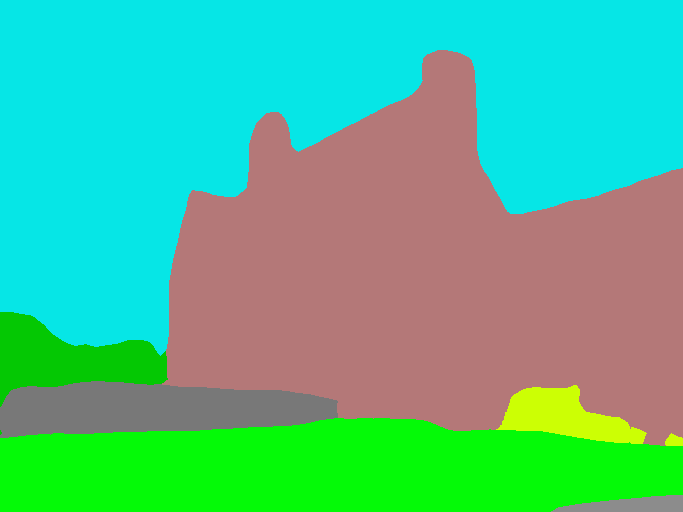}
	\end{minipage}}
	\hskip -2pt
	\subfigure{\begin{minipage}{.245\linewidth}
			\centering
			\includegraphics[width=\linewidth,height=.8\linewidth]{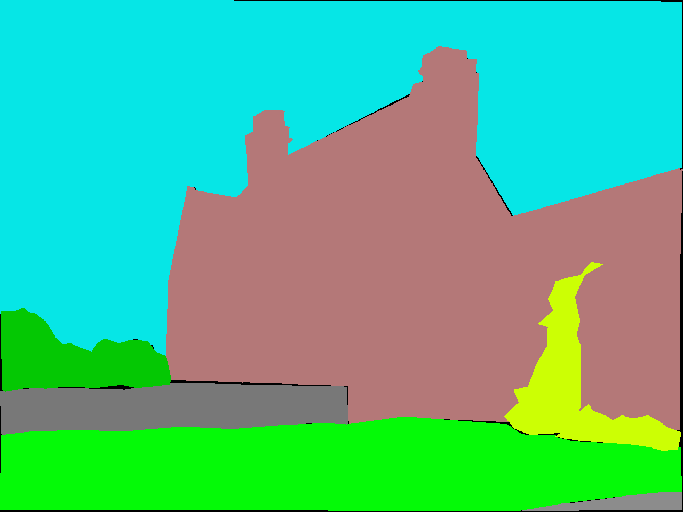}
	\end{minipage}}
	\vskip -8pt
	\subfigure{\begin{minipage}{.245\linewidth}
			\centering
			\includegraphics[width=\linewidth,height=.8\linewidth]{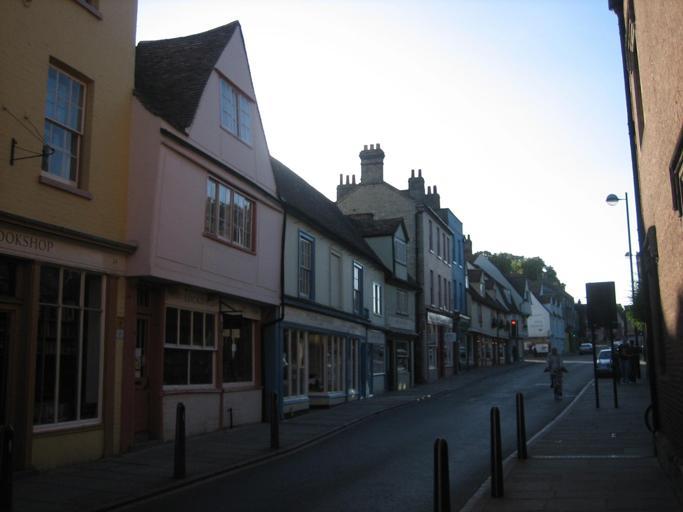}
	\end{minipage}}
	\hskip -2pt
	\subfigure{\begin{minipage}{.245\linewidth}
			\centering
			\includegraphics[width=\linewidth,height=.8\linewidth]{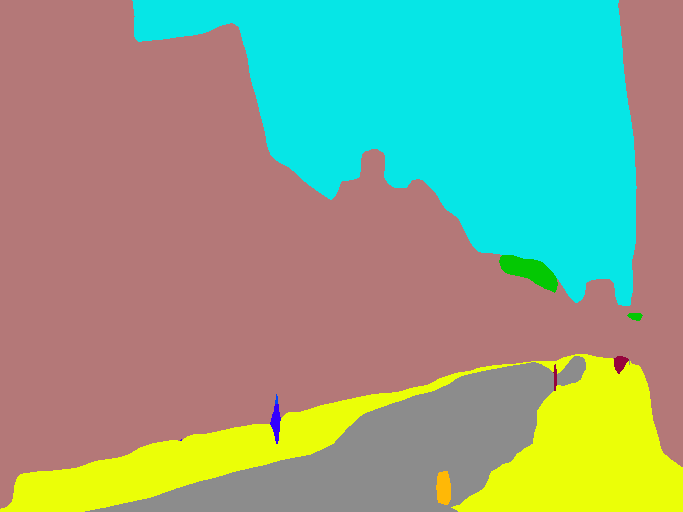}
	\end{minipage}}
	\hskip -2pt
	\subfigure{\begin{minipage}{.245\linewidth}
			\centering
			\includegraphics[width=\linewidth,height=.8\linewidth]{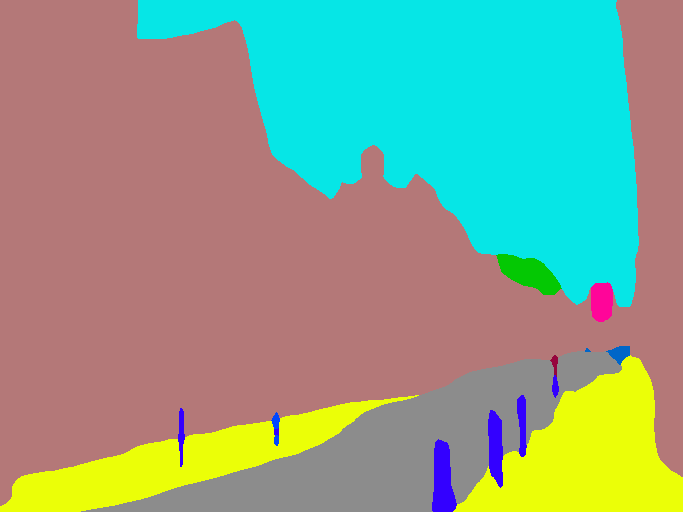}
	\end{minipage}}
	\hskip -2pt
	\subfigure{\begin{minipage}{.245\linewidth}
			\centering
			\includegraphics[width=\linewidth,height=.8\linewidth]{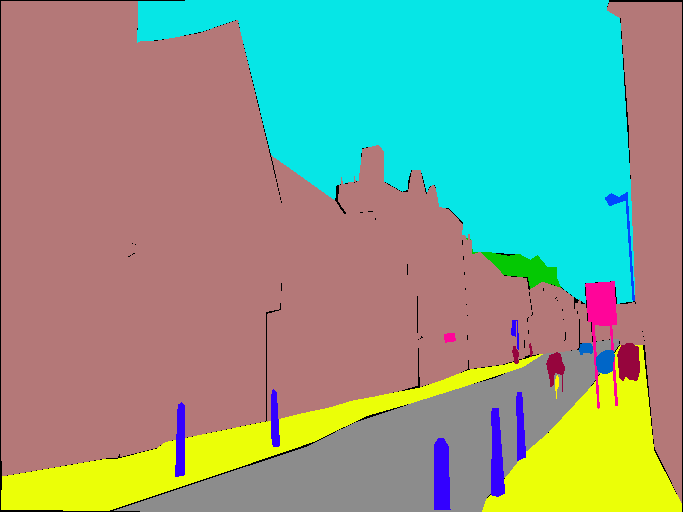}
	\end{minipage}}
	\vskip -2.5pt
	{\color{red}\rule{\linewidth}{2pt}}
	\vskip -2.5pt
	\subfigure{\begin{minipage}{.245\linewidth}
			\centering
			\includegraphics[width=\linewidth,height=.8\linewidth]{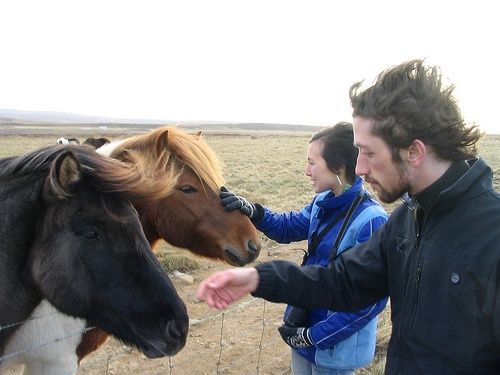}
	\end{minipage}}
	\hskip -2pt
	\subfigure{\begin{minipage}{.245\linewidth}
			\centering
			\includegraphics[width=\linewidth,height=.8\linewidth]{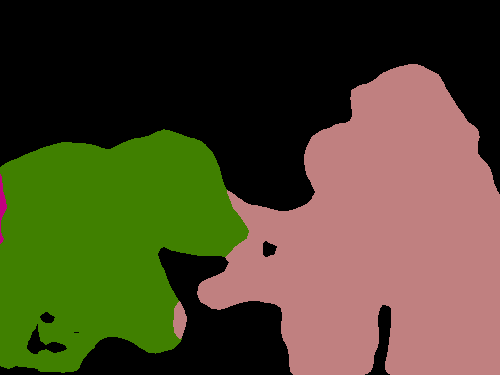}
	\end{minipage}}
	\hskip -2pt
	\subfigure{\begin{minipage}{.245\linewidth}
			\centering
			\includegraphics[width=\linewidth,height=.8\linewidth]{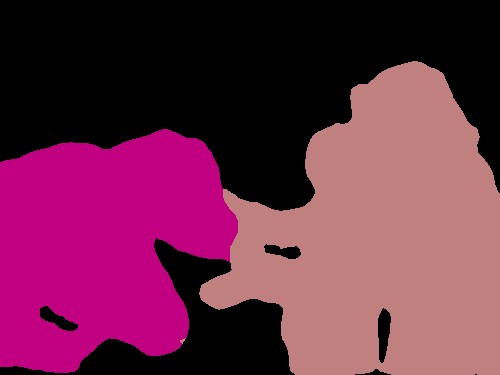}
	\end{minipage}}
	\hskip -2pt
	\subfigure{\begin{minipage}{.245\linewidth}
			\centering
			\includegraphics[width=\linewidth,height=.8\linewidth]{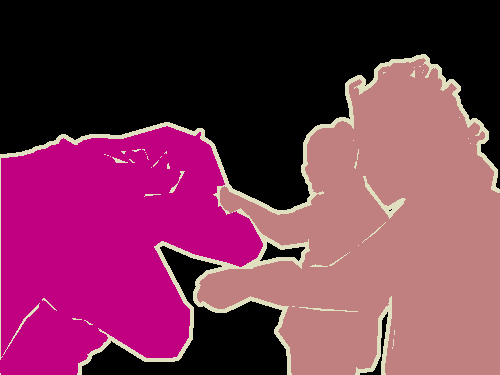}
	\end{minipage}}
	\vskip -8pt
	\subfigure{\begin{minipage}{.245\linewidth}
			\centering
			\includegraphics[width=\linewidth,height=1\linewidth]{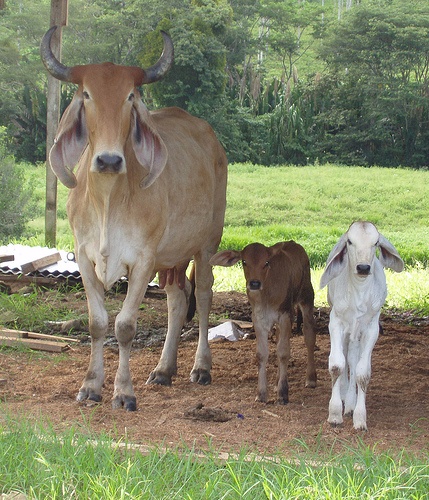}
	\end{minipage}}
	\hskip -2pt
	\subfigure{\begin{minipage}{.245\linewidth}
			\centering
			\includegraphics[width=\linewidth,height=1\linewidth]{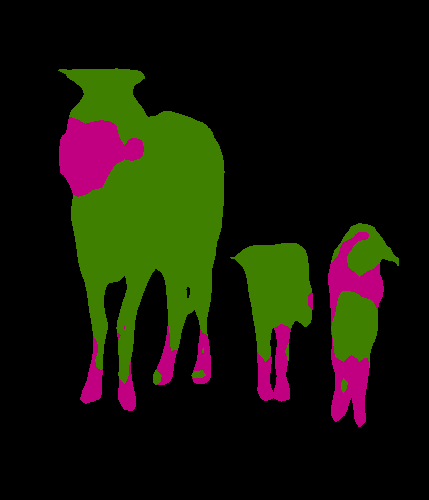}
	\end{minipage}}
	\hskip -2pt
	\subfigure{\begin{minipage}{.245\linewidth}
			\centering
			\includegraphics[width=\linewidth,height=1\linewidth]{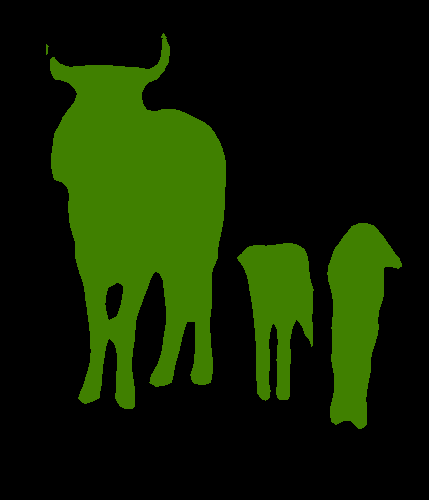}
	\end{minipage}}
	\hskip -2pt
	\subfigure{\begin{minipage}{.245\linewidth}
			\centering
			\includegraphics[width=\linewidth,height=1\linewidth]{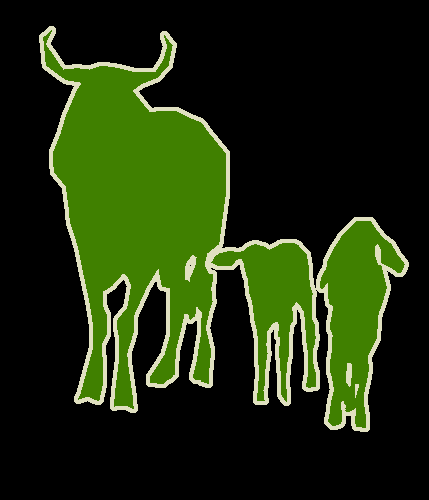}
	\end{minipage}}
	\vskip -8pt
	\setcounter{subfigure}{0}
	\subfigure[]{\begin{minipage}{.245\linewidth}
			\centering
			\includegraphics[width=\linewidth,height=1.2\linewidth]{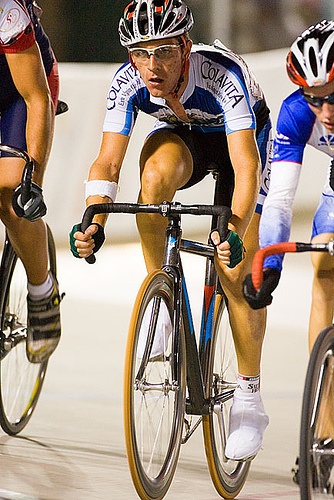}
	\end{minipage}}
	\hskip -2pt
	\subfigure[]{\begin{minipage}{.245\linewidth}
			\centering
			\includegraphics[width=\linewidth,height=1.2\linewidth]{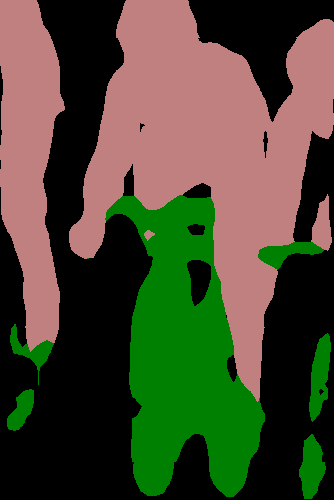}
	\end{minipage}}
	\hskip -2pt
	\subfigure[]{\begin{minipage}{.245\linewidth}
			\centering
			\includegraphics[width=\linewidth,height=1.2\linewidth]{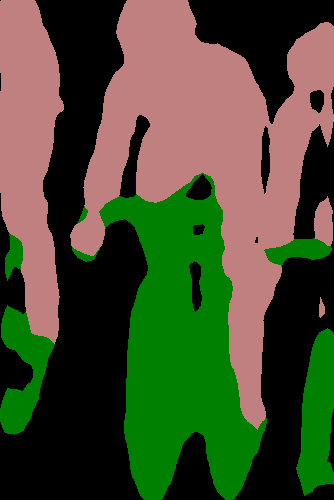}
	\end{minipage}}
	\hskip -2pt
	\subfigure[]{\begin{minipage}{.245\linewidth}
			\centering
			\includegraphics[width=\linewidth,height=1.2\linewidth]{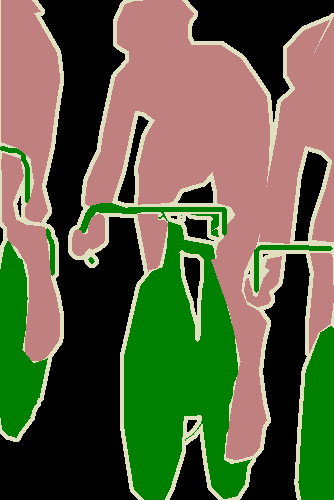}
	\end{minipage}}
	\caption{Qualitative segmentation result comparisons on Cityscapes validation set (the first three rows), ADE20K validation set (the middle three rows), and Pascal VOC 2012 validation set (the bottom three rows). The proposed DSD framework improves the performance of the student networks. (a) Input image. (b) Results of the student network. (c) Results of ours. (d) Ground truth.}
	\label{fig:results}
\end{figure}

The results on the ADE20K validation set and test set shown in Table~\ref{table:results4} also demonstrate the effectiveness and generality of our method. For MobileNetV2, the mIoU score is improved by 3.2 and 2.6 points on the validation set and test set respectively, which exceeds the accuracy of DilatedNet-50~\cite{Xiao_2018_ECCV} with large network parameters. For ResNet-18, the mIoU score is improved from 33.8\% to 38.0\% on the validation set, compared with the results 36.6\% of SKD~\cite{Liu_2019_CVPR,liu2020structured}, our method based on the same network achieves a higher segmentation result, too. On the test set, our method improves the mIoU from 27.1\% to 31.1\%, which once again proves the effectiveness and generality of our method. Finally, As shown in Fig.~\ref{fig:results}, we also visualize the segmentation results on the ADE20K validation set to illustrate the effectiveness of our method.

\section{Conclusions}
\label{conclu}

In this paper, we have presented a simple and effective framework, called double similarity distillation, to improve the classification accuracy of the compact semantic image segmentation network. Specifically, we propose the pixel-wise similarity distillation module to capture more detailed spatial dependencies across multiple layers through the residual attention maps. Besides, the category-wise similarity distillation module is proposed to strengthen the global category correlation by constructing the correlation matrix. Since there are no large amount of matrix multiplication and extra network structures, the proposed DSD framework achieves zero increase parameters and negligible FLOPs. What's more, it can be applied to all existing networks and optimized in an end-to-end manner. Finally, the ablation experiments prove the effectiveness and generality of the DSD framework. Our method achieves state-of-the-art performance on four challenging segmentation datasets. As for future work, we will consider applying our method on other dense prediction tasks.

\section*{Acknowledgment}

The authors would like to thank all the reviewers for their valuable comments and feedback.

\bibliographystyle{IEEEtran}
\bibliography{IEEEabrv,bibliography}
%









\end{document}